\documentclass[numbers]{article}

\usepackage{amsmath,amssymb,amsthm}
\usepackage{algorithmic}
\usepackage{multirow}
\usepackage{array}
\usepackage{textcomp}
\usepackage{stfloats}
\usepackage{verbatim}
\usepackage{graphicx}
\def\BibTeX{{\rm B\kern-.05em{\sc i\kern-.025em b}\kern-.08em
    T\kern-.1667em\lower.7ex\hbox{E}\kern-.125emX}}
\usepackage{balance}
\usepackage{epsfig}
\usepackage[usenames,dvipsnames]{xcolor}
\usepackage{colortbl}
\usepackage{soul}
\usepackage{makecell}
\usepackage{pifont}
\usepackage{natbib}
\usepackage{subfigure}
\usepackage{arydshln}

\usepackage[preprint]{neurips_2025}

\usepackage[utf8]{inputenc} 
\usepackage[T1]{fontenc}    
\usepackage{hyperref}       
\usepackage{url}            
\usepackage{booktabs}       
\usepackage{amsfonts}       
\usepackage{nicefrac}       
\usepackage{microtype}      
\usepackage{xcolor}         

\title{ExEBench: Benchmarking Foundation Models on Extreme Earth Events}
\author{Shan Zhao$^1$, Zhitong Xiong$^1$, Jie Zhao$^1$, Xiao Xiang Zhu$^{1,2}$\\
\\ 1. Technical University of Munich (TUM)\\ 2.  Munich Center for Machine Learning (MCML)}
\begin{document}

\maketitle

\begin{abstract}
Our planet is facing increasingly frequent extreme events, which pose major risks to human lives and ecosystems. Recent advances in machine learning (ML), especially with foundation models (FMs) trained on extensive datasets, excel in extracting features and show promise in disaster management. Nevertheless, these models often inherit biases from training data, challenging their performance over extreme values. To explore the reliability of FM in the context of extreme events, we introduce \textbf{ExE}Bench (\textbf{Ex}treme \textbf{E}arth Benchmark), a collection of seven extreme event categories across floods, wildfires, storms, tropical cyclones, extreme precipitation, heatwaves, and cold waves. The dataset features global coverage, varying data volumes, and diverse data sources with different spatial, temporal, and spectral characteristics. To broaden the real-world impact of FMs, we include multiple challenging ML tasks that are closely aligned with operational needs in extreme events detection, monitoring, and forecasting. ExEBench aims to (1) assess FM generalizability across diverse, high-impact tasks and domains, (2) promote the development of novel ML methods that benefit disaster management, and (3) offer a platform for analyzing the interactions and cascading effects of extreme events to advance our understanding of Earth system, especially under the climate change expected in the decades to come. The dataset and code are public \url{https://github.com/zhaoshan2/EarthExtreme-Bench}.
\end{abstract}

\section{Introduction}
Extreme weather and climate events, such as storms, heatwaves, floods, and droughts, are expected to become more common and more severe in the future \citep{seneviratne2021weather}. These events, exacerbated by human-induced climate change beyond natural climate variability, have caused widespread adverse impacts and related losses to nature and people \citep{parmesan2022climate}. Sequential extreme events can have cascading effects, compounding their impacts far beyond those of isolated events \citep{schauwecker2019anticipating, RN8446}. Under this circumstance, accurate prediction and detection of such events are becoming more important. These efforts not only help mitigate social and economic losses but also advance our understanding of the Earth system.

Deep learning (DL) has demonstrated promising performance in addressing both geospatial and Weather \& Climate (W\&C) related tasks \citep{zhu2024foundations}. Especially the foundation models (FMs), which are trained on vast volumes of data, excel in extracting features and can be quickly adapted to various downstream tasks. Earth Observation (EO) FMs \citep{xiong2024neural, cong2022satmae} are pre-trained on remote sensing (RS) data, such as multispectral, hyperspectral, and Synthetic Aperture Radar (SAR) imagery, and can be fine-tuned for downstream tasks like land cover mapping and flood detection. Meanwhile, some models have been specifically pre-trained on W\&C data \citep{schmude2024prithvi, nguyen2023climax}. These models have alleviated the reliance on physical models or expert knowledge while achieving improved performance regarding extended forecasting horizons, better error metrics, and uncertainty awareness. More recently, efforts have emerged to combine EO data with W\&C models \citep{schmude2024prithvi}. The unified FM holds promise to tackle equally challenging tasks across domains.

Compared to task-specific machine learning (ML) models, which are typically trained for fixed datasets and often struggle to generalize beyond their specific training conditions, FMs demonstrate advantages in handling extreme events. These extreme events are often characterized by the limited availability of labeled data and the need for rapid response. By leveraging extensive pretraining, FMs can quickly adapt to new conditions with limited training resources. 
However, despite their potential to have a significant impact on how ML is used for disaster management, FMs still need to demonstrate that they can eventually deliver better results for critical task-based applications~\citep{bauer2024if}. Like all data-driven models, FMs tend to inherit the biases from training data. Extreme events, despite their increasing intensity, duration, and spatial extent, still constitute a small portion of Earth's overall phenomena. This scarcity and imbalance presents challenges for FMs to effectively capture these events. Establishing a fair and consistent benchmark for evaluating their performance is, therefore, a non-trivial yet essential job. 

Existing benchmarks \citep{yeh2sustainbench, lacoste2024geo, bountos2023fomo} typically focus on specific domains and tasks. However, extreme events driven by interactions among numerous variables present complexities far beyond those seen in simplified, controlled benchmarks. WeatherBench \citep{rasp2020weatherbench, rasp2024weatherbench} provides a benchmark for weather forecasting models, but it does not evaluate model finetuning or transferability to other tasks. Considering the fundamental goal of FMs is to provide an efficient backbone for diverse cases, a more comprehensive platform is needed. Moreover, in the face of accelerating climate change, the ability to evaluate model adaptability to rare and extreme events is important for ensuring that these models can support critical decision-making. To address these gaps, we introduce a new benchmark in the context of extreme events, where tasks and data naturally span a wide range of resolutions, spatial extents, and modalities, to enable multidimensional performance assessment and better reflect the demands of real-world deployment.

With ExEbench, we aim to encourage the development of unified models that bridge ML, geospatial science, and W\&C community. By analyzing the interactions and cascading effects of these extreme events, we seek to enhance our understanding of the Earth system. Our main contributions are:
\begin{itemize}
    \item We release an extreme event dataset which integrates EO and weather data to evaluate the transferability of FMs across diverse extreme scenarios.
    \item We provide ready-to-use data, define ML tasks for each event, benchmark existing FMs under different fine-tuning strategies, and offer comprehensive evaluation metrics. 
    \item We outline future research directions and offer practical guidance to end users for selecting and adapting FMs to task-based applications, supporting actions against climate change.
\end{itemize}
In the following part, we will first introduce the details of ExEBench dataset and its associated tasks (Section~\ref{sec: data_task}). Then we present some experimental results on each task (Section~\ref{sec:results}), from which some key takeaways are derived.
\begin{figure*}[h!]
    \centering
    \includegraphics[width=.85\textwidth, trim={0 5.5cm 9cm 0cm},clip]{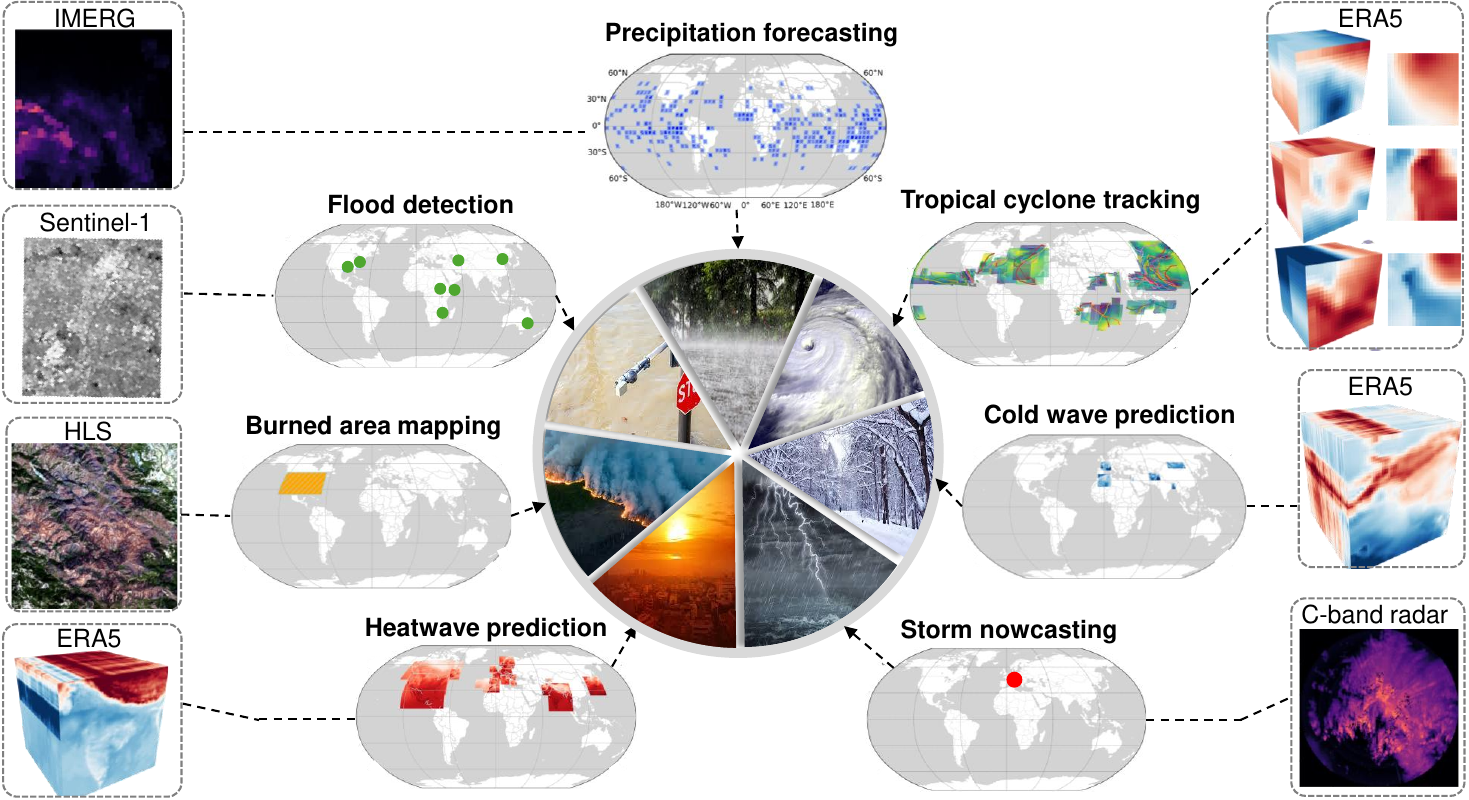}
    \caption{\small The ExEBench includes seven types of extreme events covering the global area. Various sensing techniques and Earth observation data can contribute to the detection, monitoring, and prediction of them.}
    \label{fig:global_sites}
\end{figure*}
\section{ExEBench Datasets and Tasks}\label{sec: data_task}
This section introduces the ExEBench datasets and provides background on the extreme events. In total, seven extreme event categories in Fig.\ref{fig:global_sites} are covered. We describe the data processing workflow for each event, details of the datasets, and the task defined for each event. Also, we provide standard data splits for train and test. All of the datasets are easily downloaded from Hugging Face and their data loading integrates with the PyTorch Machine Learning framework. However, we believe the scope extends well beyond the current tasks, and we encourage users to explore additional analyses.

\subsection{Overview}
The definition of extreme events is not standardized \citep{mcphillips2018defining}; in data science, extreme events are often defined as the top percentiles of observed values. However, such extreme values do not always result in socioeconomic loss. For instance, the extreme precipitation over the central Pacific might not impact human activities on land but remains crucial for understanding Earth's systems as a whole. In contrast, in disaster management, even moderate rainfall can cause substantial social impacts if the preparation is insufficient. To capture a broad perspective, we collect extreme events in two ways: one based on human-documented impacts and another on data-driven thresholds. Below, we will detail the data processing steps and explain how these events are used to evaluate FM performance across heterogeneous variables.
\subsection{Heatwaves}
\label{subsec:heatwaves}
Heatwaves (HW) are periods of exceptionally high temperatures \citep{wmo_heatwave}. Its successful early warning is essential for preparing infrastructure and mitigating the impact on agriculture \citep{ballester2023heat}. 

\begin{figure*}[h!]
    \centering
    \includegraphics[width=.8\textwidth, trim={0 3cm 6cm 0},clip]{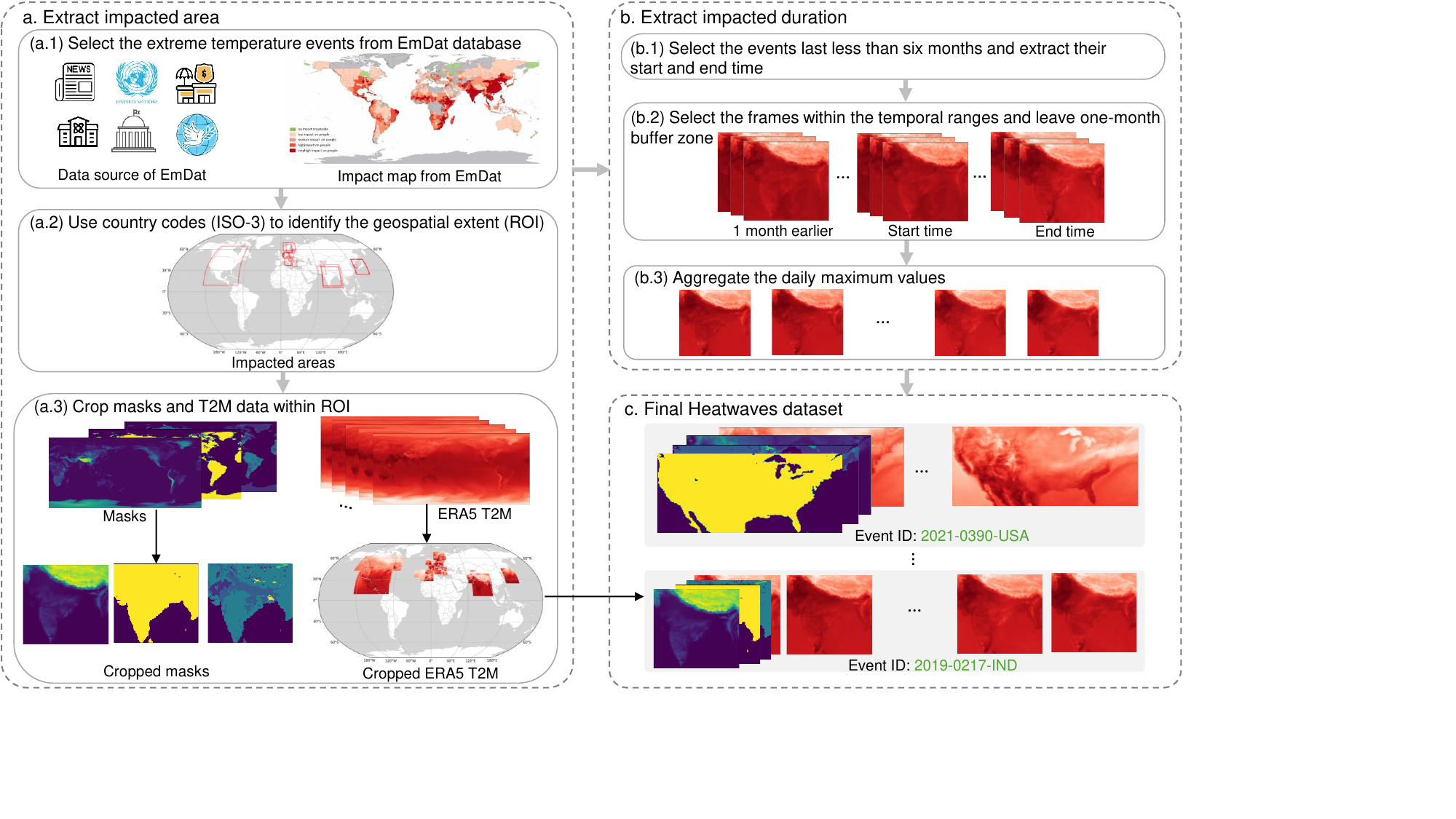}
    \caption{\small Workflow of heatwave extraction. We first identify heatwave events from EmDat database, and then prepare the corresponding local ERA5 variable during the events.}
    \label{fig:workflow_heatwave}
\end{figure*}

\textbf{Dataset} Figure~\ref{fig:workflow_heatwave} is the workflow for extracting extreme high-temperature events. First, heatwave records from 2019 to 2023 were extracted from the EmDat \citep{emdat} database. Specifically, the geographical extent and time of each heatwave were determined by the country code (ISO) and the start and end dates of the event. Heatwaves lasting less than six months were selected, with a one-month buffer added prior to each event’s start date. Second, the hourly ERA5 \citep{hersbach2020era5} temperature 2 meters ($t_{2m}$) was extracted to match the spatial and temporal extends. Last, daily maximum temperatures were aggregated to produce the final dataset. The resulting dataset contains 55 sequences, each identified by a unique event ID. These sequences vary in spatial sizes but maintain a fixed spatial resolution of 0.25$^\circ$; temporally, they range in length from 32 to 183 frames, with a daily temporal resolution. Since the original image size of the dataset varies greatly, a normalization process was applied for batched training at size $W \times W$: first, the image was resized so that the shorter side is $W$ and the aspect ratio is maintained, then a $W \times W$ region was cropped from the center of the resized image. Events from 2023 are designated as the test set, while all remaining events constitute the training set.

\textbf{Task} We define this task as a medium-range, single-step forecast. Let $X^{d}$ denote the true state of the weather state at time $d$. The time evolution of the true weather can be represented by an underlying discrete-time dynamics function, $f$, which generates the state at the next time step ($\Delta d$ in the future) based on the current one:
\begin{equation}
    \hat{X}^{d+\Delta d}_{t_{2m}} = f(X^{d}_{t_{2m}}).
\end{equation}
The $\Delta d$ is selected within a range of 0 to 14 days.

\subsection{Cold waves}
\label{subsec:coldwaves}
Cold waves (CW) relate to a period of unusually cold weather that persists for an extended duration. The data processing workflow, task definition, data transformation, and data split are similar to those used for HW, with the exception that daily \textbf{minimum} values were aggregated. Incorporating cold waves allows for examining model performance in the lower percentiles of the data. The resulting dataset contains 9 sequences.


\subsection{Storms}
\label{subsec:storms}
Precipitation is one of the most important meteorological variables. Its prediction is challenging due to its inherent stochastic nature, involvement of multi-scale processes, and complex patterns \citep{zhao2023exploring}. Most DL-based models fail to capture the intricate high-frequency features in the rain maps \citep{ayzel2020rainnet}. 

\textbf{Dataset} Figure~\ref{fig:workflow_storm} shows the workflow of extracting storms collected by a single-polarization Doppler C-Band Radar. First, daily weather reports from the weather station were reviewed to identify records with keywords like ``storm" or similar terms. Next, radar sequences from TASSRAD19 \citep{franch2020taasrad19} were filtered based on these records to compile the final dataset. The storm dataset comprises 931 sequences from 2010 to 2019, each with a spatial resolution of 500 meters, and a temporal resolution of five minutes. Events from 2019 are designated as the test set, while all remaining events constitute the training set.

\textbf{Task} We define this task as precipitation nowcasting: Given the $N-n$ historical weather states, the goal is to obtain a trajectory of $T$ future weather states:
\begin{equation}\label{eq:pcp_nowcasting}
    \hat{X}_{pcp}^{t+\Delta t: t+ T \Delta t} = f(X_{pcp}^{t-N\Delta t}, X_{pcp}^{t-(N-1)\Delta t}, ... , X_{pcp}^{t-n\Delta t}).
\end{equation}
The $T$ is set up to 25 time steps, corresponding to the nowcasting range of 0 to 2 hours. If $n=0$, the current time step is included.

\subsection{Extreme precipitation}
\label{sec:expcp}
The ultimate goal for global prediction is to resolve heavy precipitations and even cloud systems as these are responsible for the vertical energy transfer and interact with the three-dimensional large-scale circulation influencing weather globally~\citep{bauer2024if}. Unlike the storm dataset, this dataset focuses on extracting extreme precipitation (expcp) in global scale. 

\begin{figure*}[h!]
    \centering
    \includegraphics[width=.85\textwidth, trim={0 2.2cm 5cm 0cm},clip]{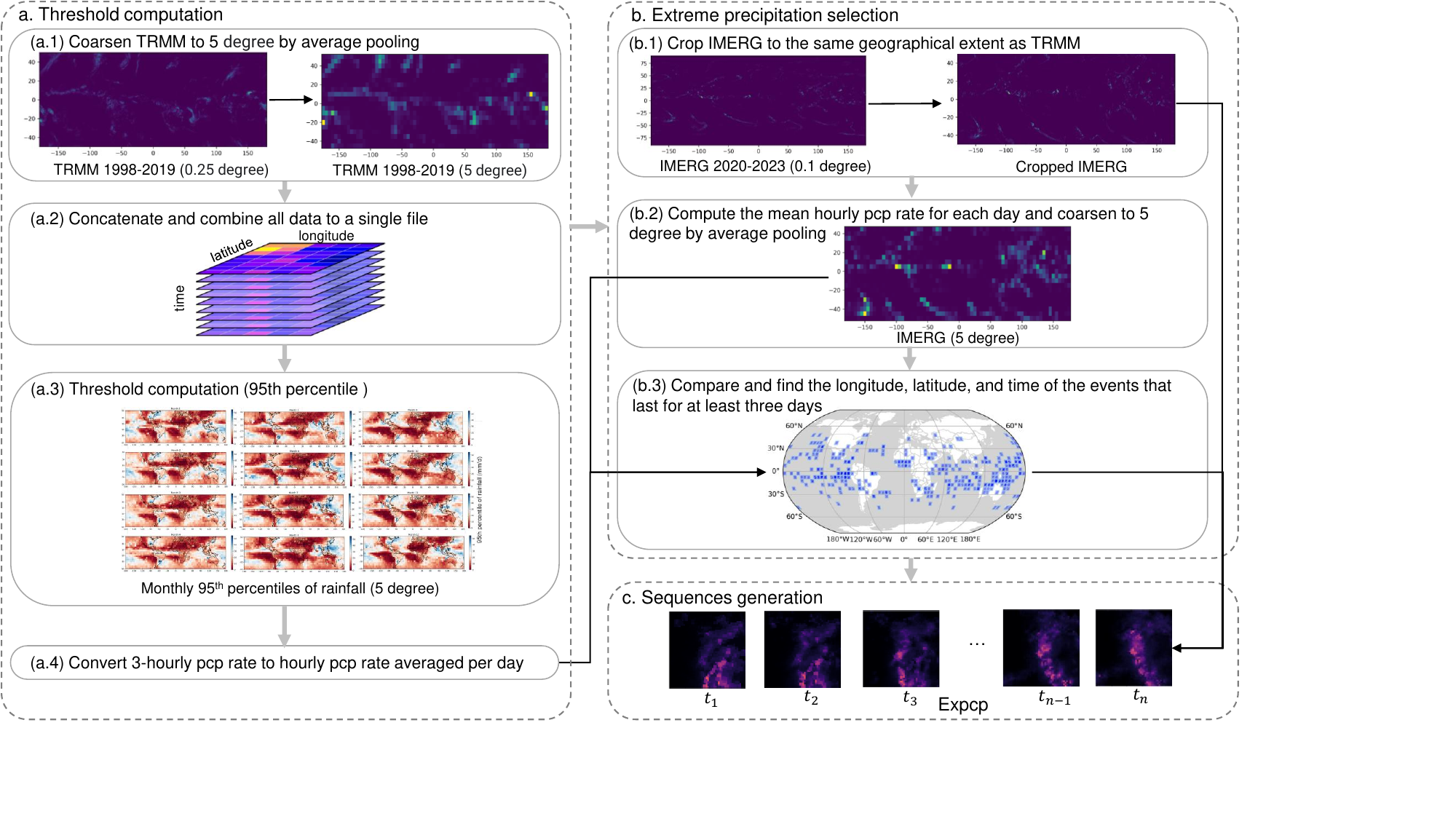}
    \caption{\small Workflow of extreme precipitation extraction. Global rainfall events exceeding the local 95th percentile values over the past 22 years are selected. (pcp: precipitation)}
    \label{fig:workflow_expcp}
\end{figure*}

\textbf{Dataset} Figure~\ref{fig:workflow_expcp} illustrates the workflow for processing precipitation variables. We used the gauge-calibrated, satellite-derived rainfall dataset TRMM 3B42 V7 \citep{huffman2007trmm}, which covers 50$^\circ$ N to 50$^\circ$ S from 1998 to 2019. First, the TRMM dataset was coarsened to a 5$^\circ$ resolution through average pooling. Monthly thresholds at each grid cell were then derived at the 95th percentile \citep{boers2019complex}(Appx.Fig.\ref{fig:expcp_pr95_trmm}). Next, rainfall data from the IMERG half-hourly Final Run global dataset at 0.1$^\circ$ resolution (GPM 3IMERGHH 07) \citep{Huffman2023} was compared with these thresholds. Careful conversions were conducted to ensure consistency in spatial and temporal dimensions and units between the two data sources (Appx.~\ref{app:data_expcp}). Consecutive periods longer than three days with rainfall above the threshold are identified as single events. The resulting dataset consists of 1092 sequences from 2020 to 2023, each with a $50\times 50$ spatial size at 0.1$^\circ$ resolution and a temporal resolution of 30 minutes. Their spatial and temporal distribution is shown in Appx.Fig.~\ref{fig:expcp_location}. Events from 2023 are designated as the test set, while all remaining events constitute the training set.

\textbf{Task} The task is precipitation nowcasting as described in Equation~\ref{eq:pcp_nowcasting}.

\subsection{Tropical cyclones}
\label{subsec:tropicalcyclones}
Tropical cyclone (TC) is the organized system of thunderstorms that develops over tropical or subtropical waters. Its formation and development is fast and involves complex and multiple meteorological variables anomalies. Recent advancements in DL for weather forecasting have shown promise in analyzing and predicting such extreme weather conditions \citep{bi2023accurate, lam2023learning, schmude2024prithvi}. 

\begin{figure*}[h!]
    \centering
    \includegraphics[width=.85\textwidth, trim={0 3.3cm 4.2cm 0cm},clip]{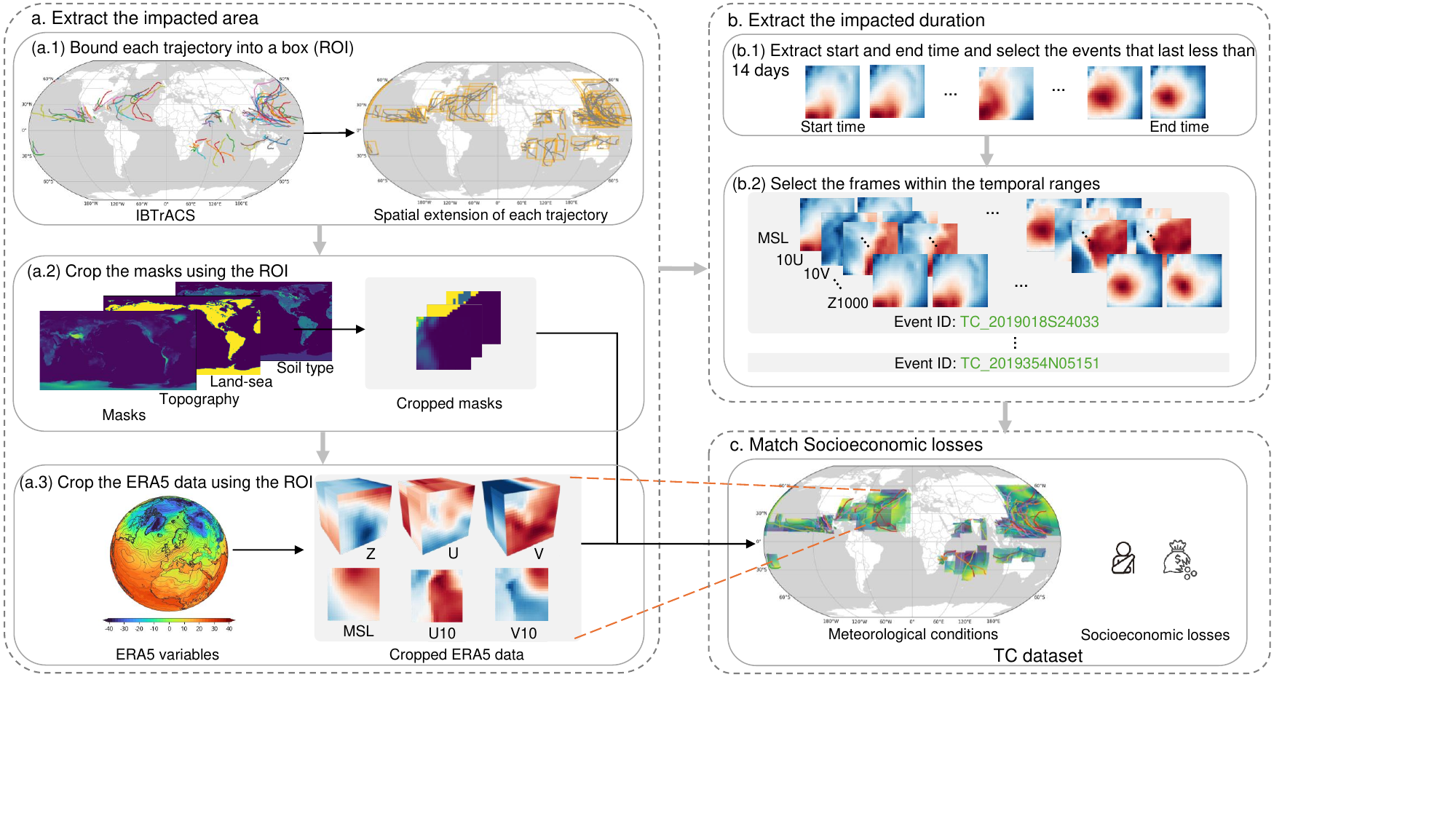}
    \caption{\small Workflow of Tropical cyclone extraction. Tropical cyclone events are selected, and the corresponding meteorological condition data during these events are prepared (a,b). Socioeconomic losses are estimated from the EmDat database (c).}
    \label{fig:workflow_TC}
\end{figure*}

\textbf{Dataset} We curated data from 95 tropical cyclone events in 2019 sourced from the International Best Track Archive for Climate Stewardship (IBTrACS) Project \citep{knapp2010international}, and prepared the corresponding meteorological conditions observed during each cyclone event as shown in Fig.~\ref{fig:workflow_TC}. First, a geographical area was defined so it bounds all trajectory points, and ERA5 variables intersecting the cyclone perimeters were extracted based on the start and end dates of the events. Only events lasting less than 14 days were considered. Each event contains three surface-level variables and three atmospheric variables that are commonly used in TC tracking algorithms. To facilitate economic and social analyses, loss information for the cyclone events was also provided by matching them with the TC records in the EmDat dataset. The temporal resolution is one hour, and the spatial resolution is 0.25$^\circ$. Data transformation follows the same method as that used in the HW dataset. Events happening later than 27th Oct, 2019 are designated as the test set, while all remaining events constitute the training set.

\textbf{Task} The multi-variable forecast is designed to predict future weather conditions based on the current weather states $X_{atm}, X_{sur}$. The objective is to generate accurate forecasts at time step $t+\Delta t$.
\begin{equation}
    \hat{X}^{t+\Delta t}_{atm}, \hat{X}^{t+\Delta t}_{sur} = f(X^{d}_{atm}, X^{d}_{sur}).
\end{equation}
Considering the rapid evolution of TC events, $\Delta t$ is set as 6 hours.

\subsection{Fires}
\label{subsec:fires}
Fires, often sparked by extreme weather conditions like dryness, high temperatures, and lightning strikes, can dramatically alter the spectral signatures of land cover \citep{kondylatos2022wildfire}. Consequently, multispectral images are frequently employed for mapping burned areas. Accurate mapping is essential for understanding fire behavior, assessing its impact on local ecology, and planning for recovery efforts. 

\textbf{Dataset} The dataset is from HLS\_burn\_scars \citep{HLS_Foundation_2023}, which contains Harmonized Landsat and Sentinel-2 imagery of burn scars and the associated masks for the years 2018-2021 over the contiguous United States. The dataset comprises 804 scenes, each with a spatial resolution of 30 meters. The data has six channels. The masks have three values: -1 for missing data, 0 for non-burned areas, and 1 for burn scars. The training set contains 540 images, and the test set has 264 images.

\textbf{Task} The task is to produce a binary burned scars map $\hat{Y} \in \mathbb{R}^{2\times 512\times 512}$ from the inputs $X \in \mathbb{R}^{6 \times 512\times 512}$.

\subsection{Floods}
\label{subsec:floods}
Urban floods, compared to open-area flood events, cause substantial social and natural damage. The accurate detection of urban floods is crucial for assessing damaged areas, planning urban water drainage systems, and prioritizing rescue efforts. SAR images are valuable resources for detecting and mapping flooded areas due to their ability to penetrate clouds. However, the urban environment is typically complex, with many objects exhibiting similar reflectivity characteristics as waters, such as double-bounce effects from trees and buildings \citep{zhao2024urban}. 

\begin{figure}[htbp]
    \centering
    \begin{minipage}[b]{0.45\textwidth}
        \centering
        \includegraphics[width=1\textwidth, trim={0 8.4cm 16cm 0cm},clip]{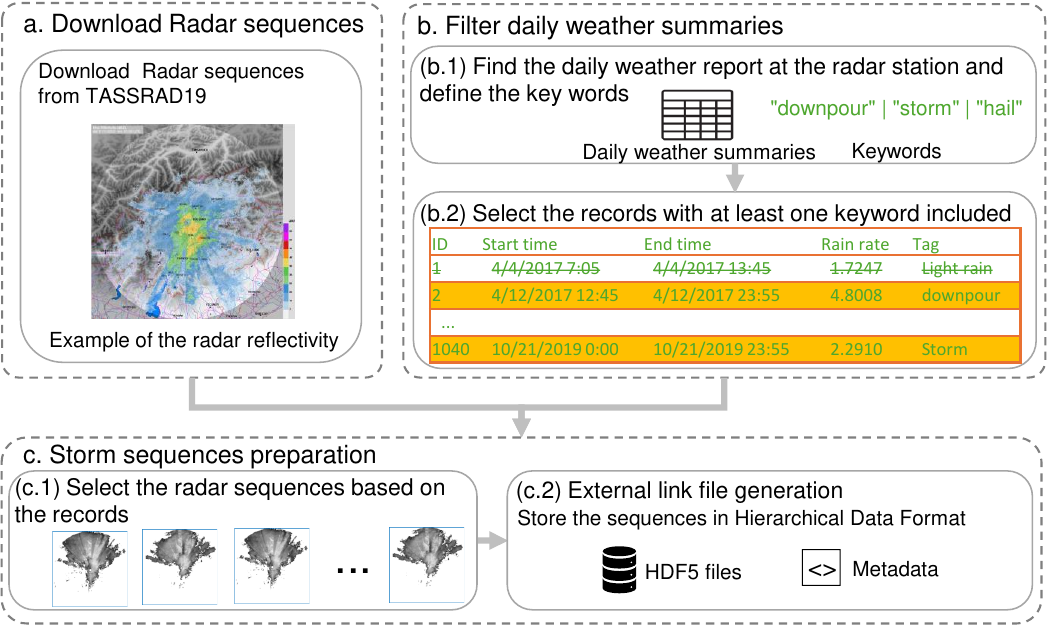}
    \caption{\small Workflow of Storms: Radar sequences are filtered based on daily weather summaries containing keywords such as ``storm''.}
        \label{fig:workflow_storm}
    \end{minipage}
    \hfill
    \begin{minipage}[b]{0.45\textwidth}
        \centering
        \includegraphics[width=.75\textwidth, trim={0 8.5cm 22cm 0cm},clip]{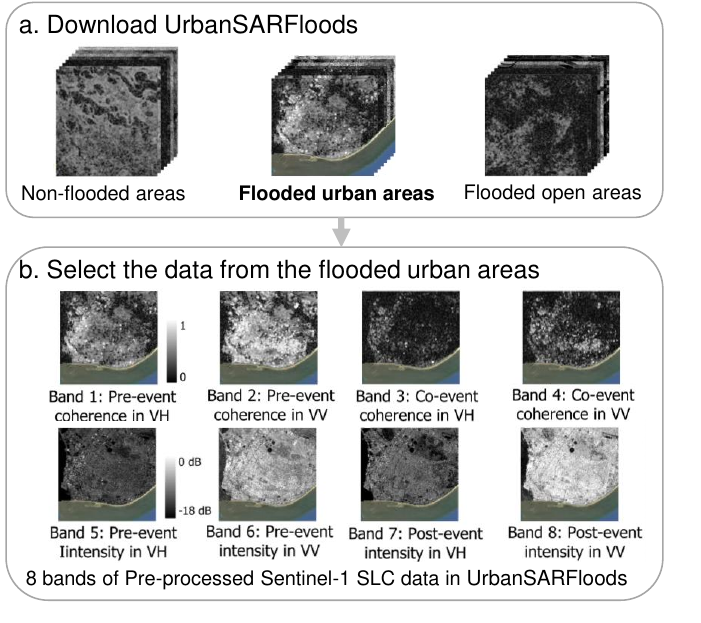}
        \caption{\small Workflow of Flood: The radar data and flood maps from flooded urban areas are selected.}
        \label{fig:workflow_flood}
    \end{minipage}
\end{figure}

\textbf{Dataset} We curated a subset from UrbanSARFloods~\citep{zhao2024urbansarfloods}, ensuring each tile contains at least one urban flooded sample, as Fig.~\ref{fig:workflow_flood} shows. The dataset contains 11 Sentinel-1 of floods and the associated masks for the years 2016-2023 over the globe, each with a size of $512\times 512$ pixels and a spatial resolution of 20 meters. The data has 8 bands, including VV and VH intensities from pre- and post-event, as well as pre- and co-event coherence. The masks have the value 0 (non-flooded), 1 (open-flooded), and 2 (urban flooded). The data split ensures geographic independence, with 405 images for training and 285 images for testing.

\textbf{Task} The task is to segment the input $X \in \mathbb{R}^{6 \times 512\times 512}$ to three classes $\hat{Y} \in \mathbb{R}^{3\times 512\times 512}$.

\subsection{Characteristics of the dataset}
This part summarizes the obstacles the dataset could pose to ML models. Addressing these challenges paves the way for enhancing model performance and deploying models in real-world scenarios.

The first problem is the distribution shift. As most FMs are pre-trained on decades of historical data, the extreme conditions are only a few part of them. Appx.Fig.\ref{fig:statistics_era5} shows the statistics of the long climatology and that of our dataset. It's observed that the mean value has shifts and the standard deviation is notably smaller.

The second challenge is data imbalance. Although the data is collected during extreme events, the majority still contains no target pixels. Appx.Fig.\ref{fig:labels_imbalance} illustrates the distribution of pixel counts for each class in the Fire and Flood dataset. For instance, the number of non-burned areas is approximately eight times greater than the number of burn scar pixels. The model may not be able to capture meaningful information, as it is biased toward the more abundant class.

\begin{figure*}[tbp!]
    \centering
    \includegraphics[width=1\textwidth, trim={0 14.6cm 0.5cm 0},clip]{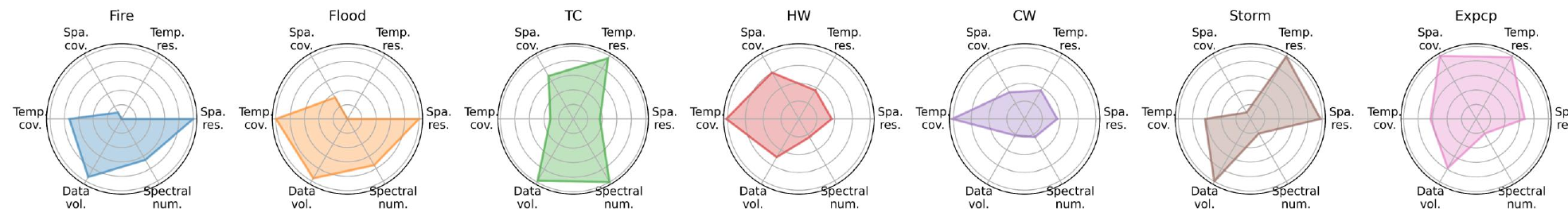}
    \caption{\small Datasets in our benchmark exhibit diverse characteristics across multiple dimensions. As the radial axis extends outward, the resolution increases or the quantity becomes higher.}
    \label{fig:data_characteristics}
\end{figure*}

Third, the data has heterogeneity in many dimensions, as Fig.\ref{fig:data_characteristics} shows. They cover regional and global phenomena, distinct in spatial granularity, span from 2010 to 2023, vary in data volumes, include both real observation and reanalysis products, and differ a lot in the spectral domain. 


More details of the datasets are in Table~\ref{tab:overview}. Considering some models require additional geographical context as input, we included meta-information for each dataset as attributes of the \texttt{Dataset} class. This meta-information includes details such as the longitudinal and latitudinal extent, start and end time, spatial resolution, and spectral bands. 

\begin{table*}[h!]
    \centering
    \resizebox{\textwidth}{!}{%
    \begin{tabular}{l p{1.6cm} p{1.7cm} p{3cm} p{2.6cm} p{2cm} p{2cm} p{2.8cm}}
         \toprule
          \textbf{Category} & \textbf{Heatwaves} & \textbf{Cold waves} & \textbf{Tropical cyclones}  & \textbf{Extreme precipitation} & \textbf{Storms} & \textbf{Fires} & \textbf{Flood}\\ \midrule
        \textbf{Data type} & \multicolumn{3}{c}{Weather} & \multicolumn{4}{c}{EO}\\ \midrule
        
        \textbf{Data source} & EmDat, ERA5, ISO-3 & EmDat, ERA5, ISO-3 & EmDat, ERA5, IBTrACSv04 & TRMM 3B42 V7, IMERG half-hourly Final Run&  TASSRAD19  &  HLS Burn Scars &  UrbanSAR- Floods\\ 
         \textbf{Sensor/variable} & $t_{2m}$ (maximum), land, soil type, topo mask & $t_{2m}$ (minimum), land, soil type, topo mask &  mslp, u10, v10; z, u and v at 1000, 850, 700, 500, 200 $hPa$, land, soil type, topo mask & Precipitation Radar and TRMM Microwave Imager& Radar precipitation rate, noise mask  & Harmonized Landsat and Sentinel-2 imagery, burned area masks & Sentinel-1, flood masks\\ \midrule
         \textbf{Spatial resolution} & \multicolumn{2}{c}{0.25$^{\circ}$} & 0.25$^{\circ}$ & 0.1$^{\circ}$ & 500 m & 30 m &  20 m \\
         \textbf{Temporal resolution} &  \multicolumn{2}{c}{Daily} & Hourly &  Half-hourly & 5-minute  & N.A &  N.A \\ 
         \textbf{Spatial coverage} &  \multicolumn{2}{c}{Global} &  Tropics& Global & Trentino South Tyrol & Contiguous United States & Global \\
         \textbf{Temporal coverage} &  \multicolumn{2}{c}{2019-2023} & 2019& 2020-2023 & 2010-2019 & 2018-2021 & 2016-2023\\ \midrule
         \textbf{Event number} & 55 &  9 & 95 &  1092 & 931 & N.A & 11 \\
         \textbf{Frame size} &  \multicolumn{2}{c}{CHW (C=4)} & CHW and CDHW (C=3, D=5) & LCHW (C=1, H,W=50,50) & LCHW (C=1, H,W=480,480) & CHW (6$\times$512$\times$512) & CHW (8$\times$512$\times$512)\\
         \textbf{Train split} & 4,844 & 338 & 12,993 & 796  &  810& 540 & 405 \\
         \textbf{Test split} & 366 frames &  221 frames & 2,438 frames & 296 sequences & 121 sequences & 264 pairs & 285 pairs\\ \midrule
         \textbf{Task} &  \multicolumn{2}{c}{Image-image prediction} & Trajectory tracking &  \multicolumn{2}{c}{Video prediction} & Segmentation & Change detection\\ 
         \textbf{Evaluation} & \multicolumn{2}{c}{ACC, RMSE, nRMSE, HQE} & ACC, RMSE & \multicolumn{2}{c}{POD, FAR, CSI, HSS} & F-1, IoU & mF-1, mIoU\\
         \bottomrule
    \end{tabular}
    }
        \caption{\small The overview of the Earth Extreme Dataset. We aim to provide data from diverse sources while ensuring the relevance of the tasks and evaluations. (topo: topography)
    }
    \label{tab:overview}
\end{table*}

\section{Results for Baseline Models}\label{sec:results}
We evaluate FMs from multiple aspects: spatial (spatial granularity, structural correctness), temporal (forecasting horizons), efficiency (parameter efficient fine-tuning), spectral (spectral bands/variables the model can handle), and task diversity (regression, segmentation, or forecasting).

The model performance was evaluated using a set of comprehensive metrics, including normalized RMSE (nRMSE), anomaly correlation coefficient (ACC), relative quantile errors (RQE) \citep{kurth2023fourcastnet}, Probability of Detection (POD), Critical Success Index (CSI), Heidke-Skill-Score (HSS), F1-score, and Intersection over Union (IoU), etc. The detailed metrics used in each task are in the Appx.~\ref{app:evaluation}. 

\begin{figure}[bp!]
    \centering
    \includegraphics[width=.93\textwidth, trim={0 4.3cm 0cm 0},clip]{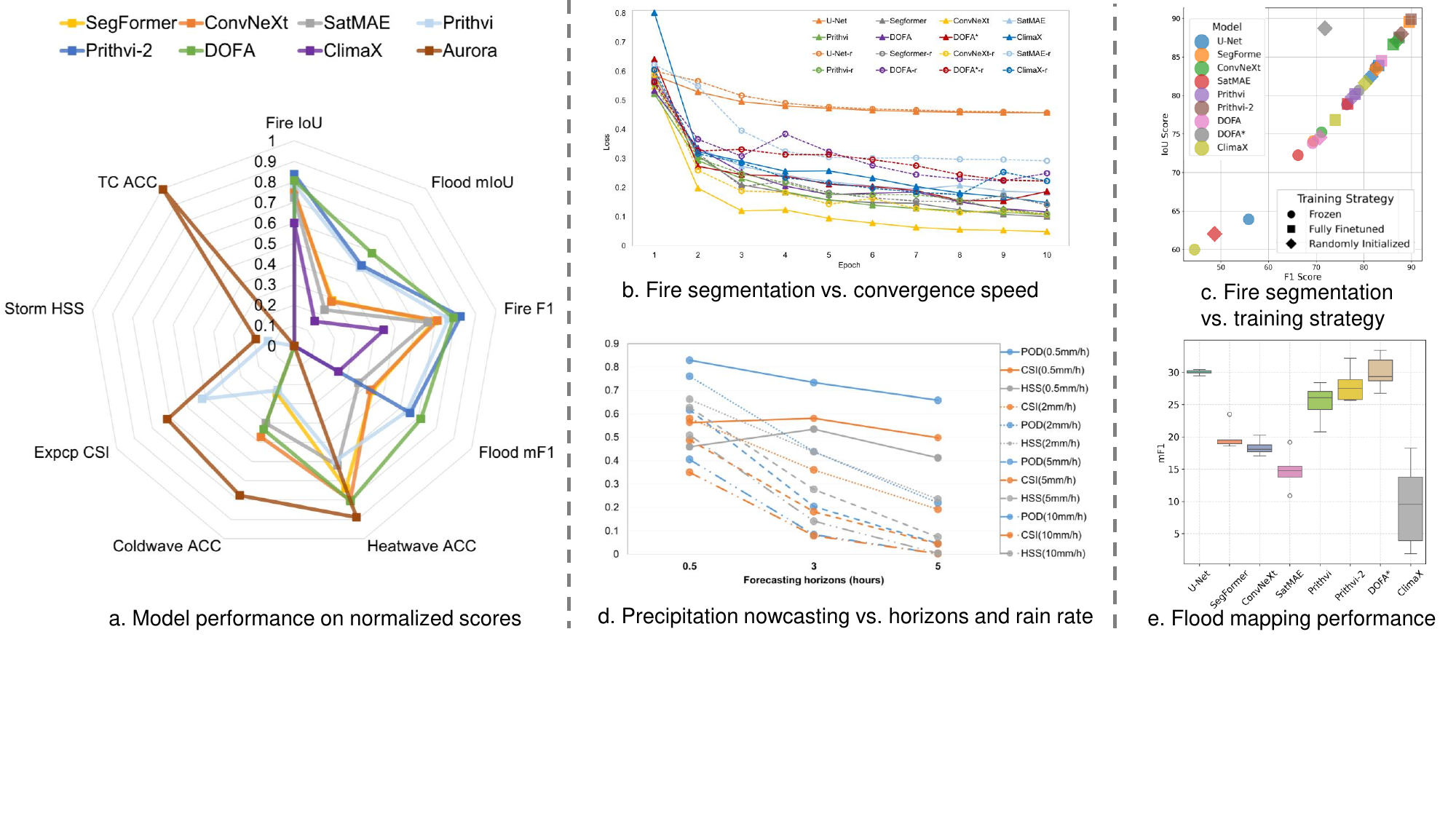}
    \caption{\small Sample results. (a) Model performance on normalized scores. A score of 0 denotes either failure to process its data modality or complete task failure. (b) The fire segmentation using pre-trained weights ($\triangle$) and randomly initialized weights ($\circ$). (c) Fire segmentation under different training strategies. (d) Precipitation nowcasting across varying forecasting horizons. (e) Flood mapping (mean and std of five runs). Detailed numbers are in Appx.~\ref{app:experiments}.}
    \label{fig:results}
\end{figure}

We include three types of FMs based on their pre-training data: vision images (U-Net \citep{ronneberger2015u}, SegFormer \citep{xie2021segformer}, ConvNeXt \citep{liu2022convnet}), EO data (SatMAE \citep{cong2022satmae}, Prithvi \citep{jakubik2023foundation}, Prithvi-2~\citep{szwarcman2024prithvi}, DOFA \citep{xiong2024neural}), and W\&C data (ClimaX \citep{nguyen2023climax}, AURORA \citep{bodnar2024aurora}), as detailed in Table~\ref{tab:models}. By conducting experiments on datasets from W\&C and EO domains, we assess the models' cross-data-type generalizability. To compare different fine-tuning strategies, we report model performance across three approaches: freezing the model body and fine-tuning only the decoder, fully fine-tuning the model, and fine-tuning using Low-rank adaptation (LoRA) \citep{hu2021lora}. To validate the effectiveness of the pretrained data, we provide fully finetuned results from the models with random weights. The pre-trained encoder extracts features from the input data, which are subsequently processed using either a lightweight CNN-based decoder or UperNet \citep{xiao2018unified} for the output. All experiments were conducted using an NVIDIA RTX A6000 GPU with 48 GB of memory. Sample results are shown in Fig.~\ref{fig:results}. 
\begin{table*}[h!]
    \centering
    \resizebox{\textwidth}{!}{%
    \begin{tabular}{p{2.5cm}p{1.2cm}p{1cm}p{6.5cm}p{1cm}p{1cm}p{1cm}p{1cm}p{1cm}p{1cm}p{1cm}}
    \toprule
       \textbf{Model} & \textbf{Para-meters}& \textbf{Type} & \textbf{Pretrained data} & \textbf{HW} & \textbf{CW} & \textbf{TC} &  \textbf{Expcp} & \textbf{Storms} & \textbf{Fires} & \textbf{Floods} \\
    \midrule
       U-Net \citep{ronneberger2015u} & 7 M & & LGG Brain MRI~\citep{buda2019lgg} &\ding{51} & \ding{51} & \ding{55} & \ding{51} &\ding{55} & \ding{51} & \ding{51} \\ 
       SegFormer \citep{xie2021segformer}& 4 M &\multirow{2}{*}{Vision} & ImageNet& \textcolor{red}{\ding{51}} & \textcolor{red}{\ding{51}} & \textcolor{red}{\ding{55}} & \textcolor{red}{\ding{55}} & \textcolor{red}{\ding{51}} & \textcolor{red}{\ding{51}} & \textcolor{red}{\ding{51}}  \\ 
       
       ConvNeXt \citep{liu2022convnet} & 60 M & & ImageNet & \textcolor{red}{\ding{51}} & \textcolor{red}{\ding{51}} & \textcolor{red}{\ding{55}} & \textcolor{red}{\ding{55}} & \textcolor{red}{\ding{51}} & \textcolor{red}{\ding{51}} & \textcolor{red}{\ding{51}} \\  \midrule 
       
       SatMAE \citep{cong2022satmae}& 187 M &\multirow{4}{*}{EO}& fMoW Sentinel (B1-12 and B8A) & \textcolor{red}{\ding{51}} & \textcolor{red}{\ding{51}} & \textcolor{red}{\ding{55}} & \textcolor{red}{\ding{55}} & \textcolor{green}{\ding{55}} & \textcolor{green}{\ding{51}} & \textcolor{green}{\ding{51}} \\
       
       Prithvi \citep{jakubik2023foundation} & 129 M& & HLS dataset & \textcolor{red}{\ding{51}} & \textcolor{red}{\ding{51}} & \textcolor{red}{\ding{55}} & \textcolor{red}{\ding{51}} & \textcolor{green}{\ding{51}} & \textcolor{green}{\ding{51}} & \textcolor{green}{\ding{51}} \\ 
       
      Prithvi-2~\citep{szwarcman2024prithvi} & 353 M && HLS dataset, Sentinel-2 & \textcolor{red}{\ding{55}} & \textcolor{red}{\ding{55}} & \textcolor{red}{\ding{55}} & \textcolor{red}{\ding{55}} & \textcolor{green}{\ding{55}} & \textcolor{green}{\ding{51}} & \textcolor{green}{\ding{51}}   \\
       DOFA \citep{xiong2024neural} & 152 M && Sentinel-1, Sentinel-2, Gaofen, NAIP, EnMAP & \textcolor{red}{\ding{51}} & \textcolor{red}{\ding{51}} & \textcolor{red}{\ding{55}} & \textcolor{red}{\ding{55}} & \textcolor{green}{\ding{55}} & \textcolor{green}{\ding{51}} & \textcolor{green}{\ding{51}} \\\midrule
       ClimaX \citep{nguyen2023climax} & 149 M &\multirow{2}{*}{W\&C}& CMIP6 & \textcolor{green}{\ding{55}} & \textcolor{green}{\ding{55}} & \textcolor{green}{\ding{55}} & \textcolor{green}{\ding{55}} & \textcolor{red}{\ding{55}} & \textcolor{red}{\ding{51}} & \textcolor{red}{\ding{51}} \\ 
       AURORA \citep{bodnar2024aurora} & 1,256 M && ERA5, CMCC, IFS-HR, HRES Forecasts, GFS Analysis and GFS Forecasts & \textcolor{green}{\ding{51}} & \textcolor{green}{\ding{51}} & \textcolor{green}{\ding{51}} & \textcolor{green}{\ding{51}} & \textcolor{red}{\ding{51}} & \textcolor{red}{\ding{55}} & \textcolor{red}{\ding{55}} \\ 
    \bottomrule
    \end{tabular}%
    }
    \caption{\small Foundation Models and Experiments: The check marks indicate all experiments conducted in the paper using the foundation model corresponding to each row. The \textcolor{red}{red} color indicates cross-data-type evaluation (from EO to W\&C and vice versa, from vision to EO, from vision to W\&C), while the \textcolor{green}{green} color denotes in-data-type experiments (U-Net is randomly initialized).}
    \label{tab:models}
\end{table*}

Overall, the pre-trained weights show benefits in faster convergence and improved performance in handling extreme events, especially when the downstream task shares the same data modality and source as the pre-trained model (Fig.\ref{fig:results} a-c). Key takeaways from our baseline experiments include: 
    
    \noindent \textbf{Temporal dynamics} FMs have large room for improvement in capturing temporal dynamics. Most FMs either can't process temporal data or suffer degraded performance with long horizons (Fig.\ref{fig:results} d).

    \noindent \textbf{Model design.} RS and W\&C data processing necessitate specialized considerations during model design. Future models should consider the spectral, geographical, and physical properties of the variables. For example, the improved performance of Prithvi-2 over Prithvi highlights the benefits of using location embeddings. However, many models still overlook critical data-specific features, such as radar signal variations across polarization modes, which can lead to failed mapping (Fig.\ref{fig:results} e). 
    
    \noindent \textbf{Transfer to limited labels} Given the often limited data available for fine-tuning, incorporating transfer learning and regularization techniques can be advantageous. Future FMs could explore knowledge distillation from complex models to simpler architectures to handle downstream tasks with minimal labeled data.

    \noindent \textbf{Model choice} End users can select an FM whose pretrained data is closely aligns with the downstream tasks in terms of spectral, spatial, and temporal dimensions.
        
    \noindent \textbf{Synergy of geospatial and weather \& climate FMs}
    Given the improved generalizability of FMs across data modalities, there is a potential for geospatial FMs and W\&C models to be applied to equally challenging tasks. 

For more task-specific discussions, please refer to Appx.~\ref{app:experiments}.


\section{Conclusion}\label{sec:conclusion}
To conclude, we developed a benchmark with seven extreme events, and we designed ML tasks for each event type. The datasets show high diversity across spatial, temporal, and spectral domains, and the task design accounts for both data characteristics and the practical needs of end-users. The baseline experiments compared various model architectures, fine-tuning strategies, sizes of training data, and, for the first time, tested cross-data-type generalization (e.g., transferring knowledge from RS to weather data). A comprehensive evaluation protocol from both ML and domain-specific perspectives was used to assess the model's performance. 

Through extensive experimentation, we explored the effectiveness and potential of FM backbones in various transfer scenarios, such as transitioning from global to local forecasts, adapting to different spatial resolutions, adjusting forecasting horizons, transferring from vision and RS imagery to weather data, scaling from high-resolution to coarse-resolution data, modifying from slow- to rapidly changing dynamics, and shifting between segmentation and nowcasting tasks. Such diversity provides critical insights into the strengths and limitations of different FMs. Finally, we outlined a potential future scenario where geospatial and W\&C models could perform equally well across diverse and challenging extreme events. One of the key concerns is improving model generalizability. The model design should carefully consider the physical, geographical, and spectral characteristics of the Earth observation or simulation data. 

The dataset and task definition have broad societal impacts. We expect that this benchmark will encourage the development of FMs with enhanced generalizability for tackling various scenarios and extreme values. This, in turn, will pave the way for more robust disaster management strategies in the face of accelerating climate change. Accurate and timely detection of extreme events will reduce rescue efforts and minimize societal losses. However, ML models may produce imperfect predictions that influence policy decisions and lead to negative societal consequences. For example, models often under-predict extreme temperatures. Our dataset also contains location information, and the imbalanced distribution over the globe may lead to training bias towards certain areas. For instance, the flood dataset focuses on urban areas, which may cause models to overlook rural regions. If such a model were used for aid distribution, rural populations could be underrepresented and receive insufficient support.

In the future, the benchmark could be expanded to encompass an even broader spectrum of extreme events and integrate more data modalities. For instance, landslides could be monitored using unmanned aerial vehicle data fused with text data. Future evaluations could also incorporate more domain-specific metrics to ensure relevance for end users. Furthermore, incorporating baseline results from traditional physical or task-specific models would allow for a fair understanding of how ML is revolutionizing the field and how FMs can enhance task-specific applications.
\section{Acknowledgements}
The work is jointly supported by the German Federal Ministry of Education and Research (BMBF) in the framework of the international future AI lab ``AI4EO -- Artificial Intelligence for Earth Observation: Reasoning, Uncertainties, Ethics and Beyond'' (grant number: 01DD20001), by German Federal Ministry for Economic Affairs and Climate Action in the framework of the ``National Center of Excellence ML4Earth'' (grant number: 50EE2201C), by the Excellence Strategy of the Federal Government and the Länder through the TUM Innovation Network EarthCare, by Munich Center for Machine Learning, and by the German Federal Ministry for the Environment, Nature Conservation, Nuclear Safety and Consumer Protection (BMUV) based on a resolution of the German Bundestag (grant number: 67KI32002B; Acronym: EKAPEx) 
\vfill
\pagebreak
\bibliographystyle{neurips}
\bibliography{refs}
\vfill
\pagebreak
\section{Supplementary Material}
\subsection{Dataset Details}
In this section, we provide additional details of the datasets.
\paragraph{Heatwaves, cold waves, and tropical cyclones}
The heatwaves (HW) and coldwaves (CW) in our dataset are associated with the 2-meter temperature ($t_{2m}$). For tropical cyclones (TC), we include relevant surface variables such as mean sea level pressure (mslp), u-component and v-component of 10m wind (u10, v10), as well as atmospheric variables including geopotential (z), u-component and v-component of wind speed (u, v) at pressure levels 1000, 850, 700, 500, 200 $hPa$.

\begin{figure*}[htbp!]
    \centering
    \includegraphics[width=1\textwidth, trim={0 8cm 9cm 0cm},clip]{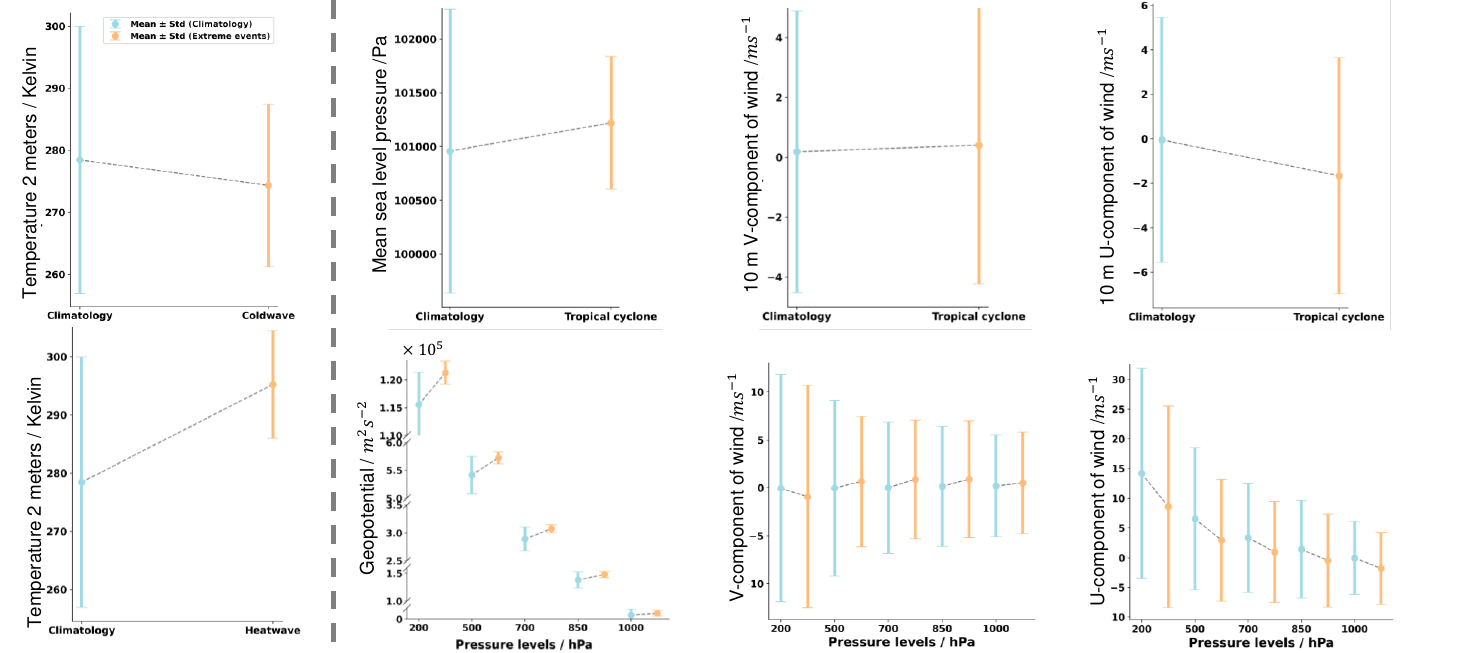}
    \caption{\small Comparison of statistics between a 39-year long-term climatology and extreme events (Heatwaves, cold waves, and tropical cyclones). The mean values observed during extreme events generally show a significant deviation from the long-term climatological averages. However, the variability (or standard deviation) during these extreme events is typically smaller compared to that of the long-term climatology.}
    \label{fig:statistics_era5}
\end{figure*}

To facilitate model training and evaluation, we provide the mean and standard deviation of each dataset for normalization purposes. Figure~\ref{fig:statistics_era5} compares the statistics of the HW, CW, and TC datasets and long-term climatology. The long-term climatology of ERA5 data is downloaded from WeatherBench2~\citep{rasp2024weatherbench}.

\begin{figure}[htbp!]
    \centering
    \includegraphics[width=1\textwidth]{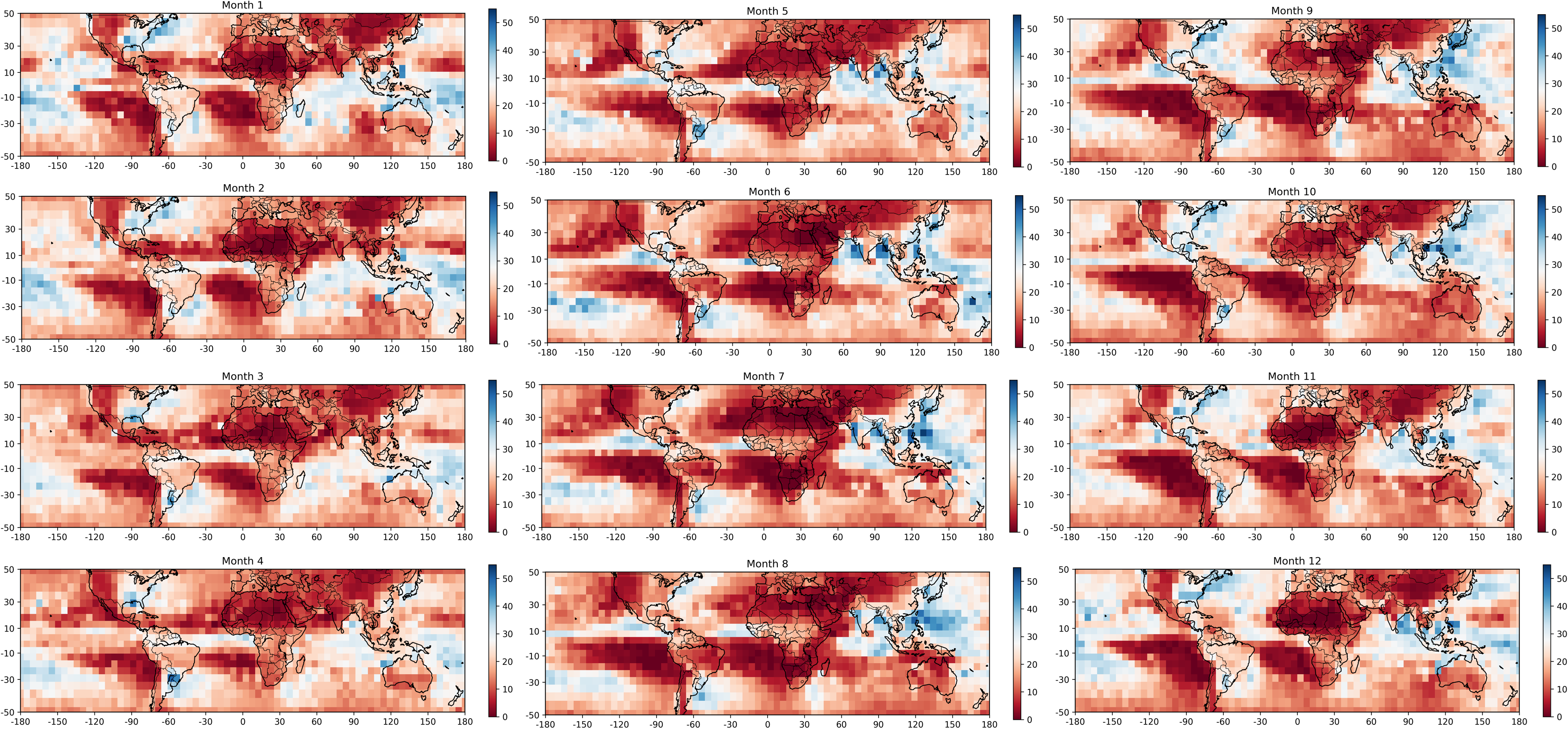}
    \caption{\small Extreme precipitation: The global monthly 95 percentile precipitation rate (mm/d) from 22-year TRMM dataset. Every grid point has its own extreme precipitation threshold, depending on the climatology.}
    \label{fig:expcp_pr95_trmm}
\end{figure}

\paragraph{Extreme precipitation}\label{app:data_expcp}
The monthly 95th percentile values of global precipitation (RP95) is computed from the Tropical Rainfall Measuring Mission (TRMM) data as follows: For each grid point (defined by latitude and longitude) in the coarsened TRMM dataset ($5^\circ\times5^\circ$ resolution), we iterate through each month of the year (January to December). For a given month and grid point, we aggregate all precipitation records from the past 22 years, sort them in ascending order, and compute the 95th percentile. This value serves as the precipitation threshold exceeded by the top 5\% of values for that specific location and month. Figure~\ref{fig:expcp_pr95_trmm} shows the resulting RP95 estimates over the past two decades. Generally speaking, RP95 values in tropical regions are larger than those in mid-high latitudes. These monthly thresholds are subsequently used to filter extreme precipitation events.

\begin{figure}[htbp!]
    \centering
    \includegraphics[width=1\textwidth, trim={0 2.6cm 0cm 9.5cm},clip]{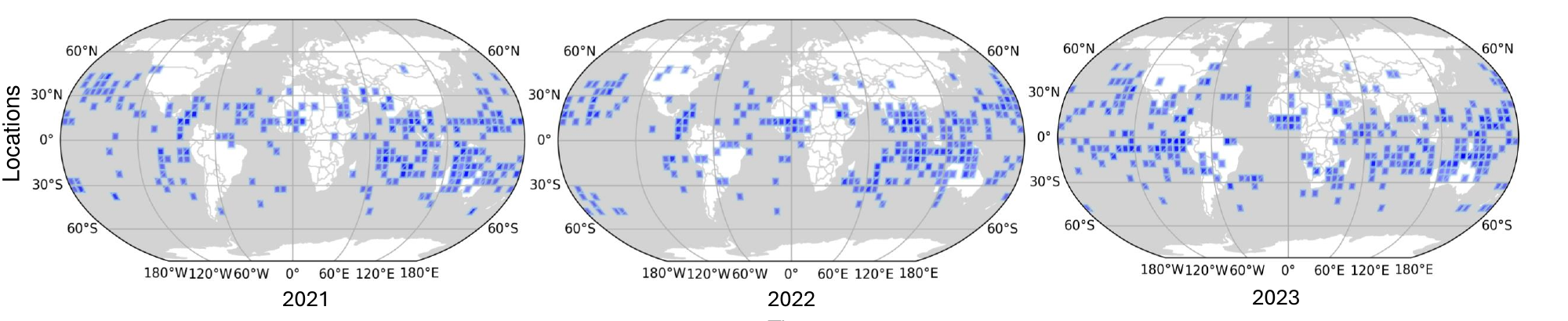}
    \caption{\small Extreme precipitation: The locations of events in the final Expcp dataset.}
    \label{fig:expcp_location}
\end{figure}

Figure~\ref{fig:expcp_location} displays the geographic locations of the extreme precipitation events for the years 2021, 2022, and 2023. The events span latitudes from 50$^\circ$N to 50$^\circ$S.

\paragraph{Fire and Flood}
The fire dataset includes six spectral bands: Blue (B02), Green (B03), Red (B04), Near-Infrared (NIR, B8A), Shortwave Infrared 1 (SWIR1, B11), and Shortwave Infrared 2 (SWIR2, B12).

The flood dataset comprises eight bands derived from Sentinel-1 observations: two bands (VV and VH) for Sentinel-1 intensity acquired pre-event, two bands (VV and VH) for Sentinel-1 intensity acquired post-event, two bands (VV and VH) for Sentinel-1 coherence acquired preevent, and two bands (VV and VH) for Sentinel-1 intensity acquired co-event. The data split ensures geographic independence, with training data covering flood events in Beira, Iran, Canada, Japan, Port Macquarie, Sydney, Coraki Niger, Hebei, Beledweyne, and the test data covers events from Houston and Lumberton.

\begin{figure}[htbp!]
    \centering
    \subfigure[Fire]{%
        \includegraphics[width=0.45\textwidth, trim={0 1cm 0 0cm},clip]{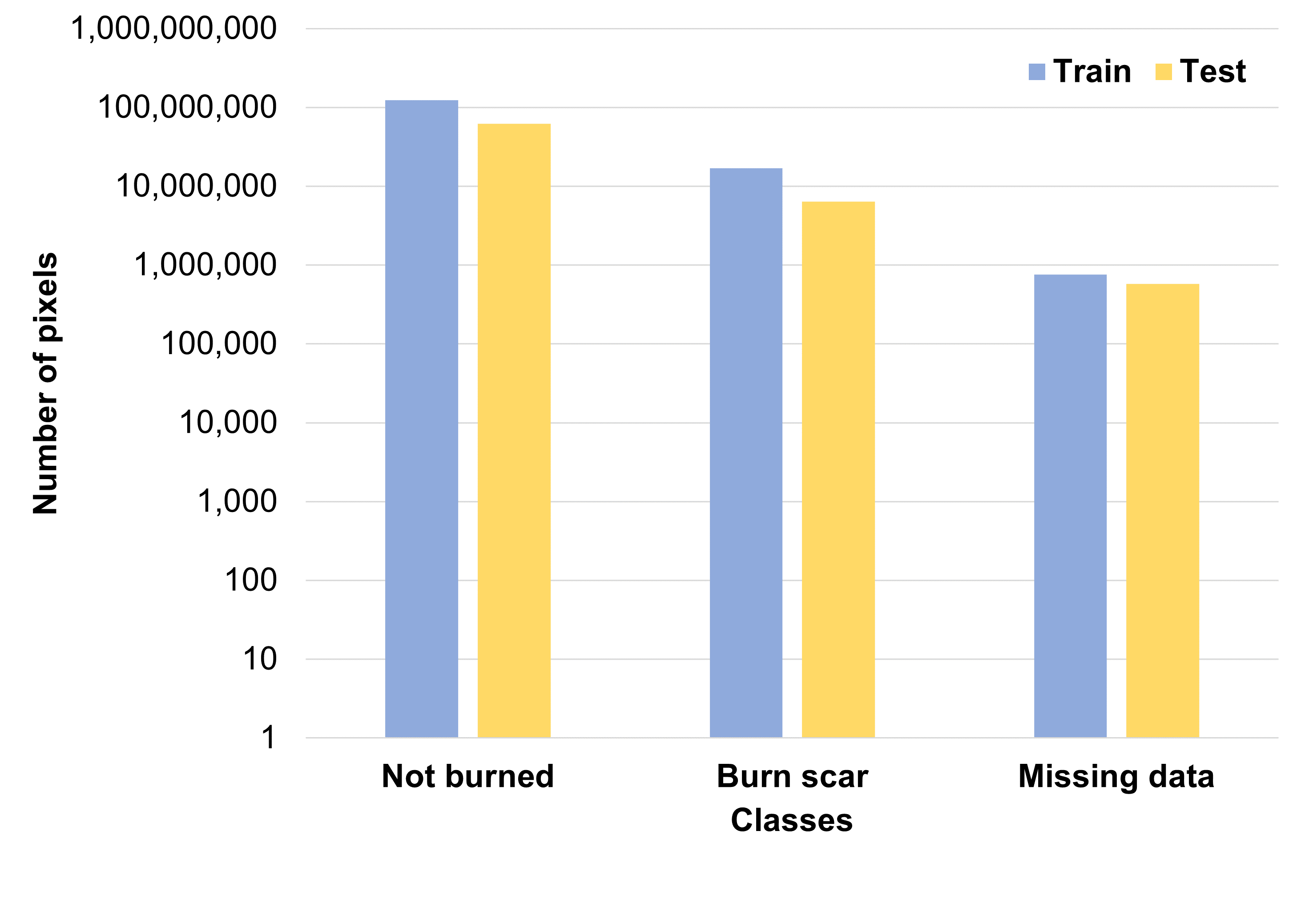}}
    \hfill
    \subfigure[Flood]{%
        \includegraphics[width=0.45\textwidth, trim={0 1cm 0 0cm},clip]{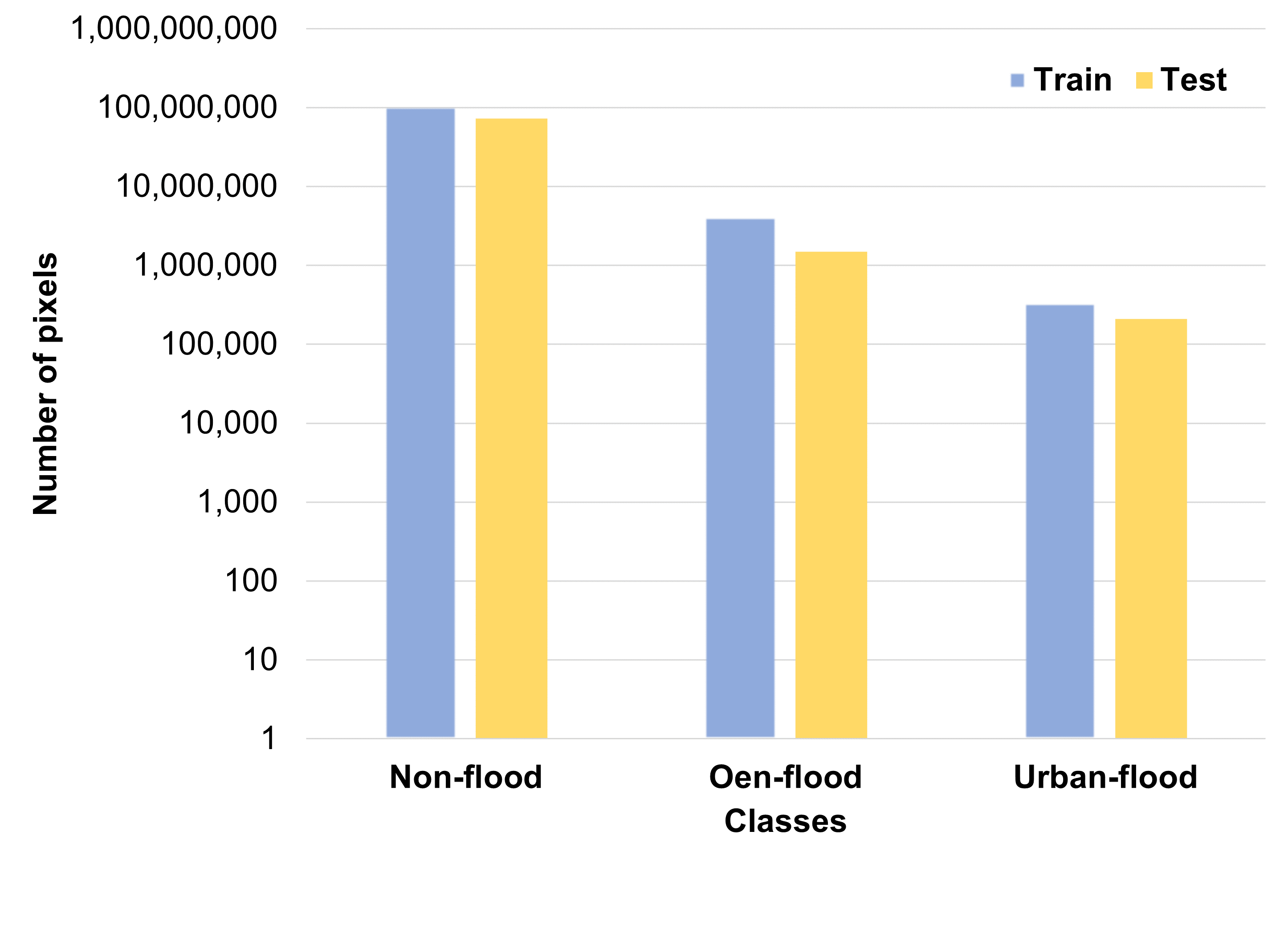}}
        \caption{\small The distribution of the labels in the Fire and Flood datasets. We use a log-scaled $y$-axis. The number of pixels with targets (burned scars and flood) is much lower than the background pixels.}
    \label{fig:labels_imbalance}
\end{figure}

Figure~\ref{fig:labels_imbalance} visualizes the label distribution in the Fire and Flood datasets. Both exhibit label imbalance, a common challenge in many geospatial tasks.

\subsubsection{Dataset Licenses}\label{app:licence}
The datasets included in ExEBench are governed by various licenses to support lawful and open use. Landsat satellite images are in the public domain, allowing unrestricted use, reproduction, and distribution. Similarly, Sentinel-1 imagery is freely accessible with permissions for reproduction and distribution. The EmDat database grants free access to non-commercial institutions for research purposes. Copernicus ERA5 data permits lawful use across a wide range of purposes, including reproduction, distribution, public communication, adaptation, modification, and integration with other data. The TAASRAD19, UrbanSARFloods radar scans, and IMERG precipitation products are released under the Creative Commons Attribution 4.0 International license; TRMM data is publicly available without usage restrictions.

ExEBench itself is released under a CC BY 4.0 license, ensuring compatibility with all included datasets.

\subsection{Evaluation metrics }\label{app:evaluation}
\subsubsection{Heatwaves and cold waves}
We use Root Mean Square Error (RMSE), normalized RMSE (nRMSE), and Anomaly Correlation Coefficient (ACC), defined below, to evaluate model performance. For a given time point $t$
\begin{equation}
    \text{RMSE} = \sqrt{\frac{1}{N} \sum_{i=1}^{N} (\hat{x}_i - x_i)^2},
\end{equation}
and 
\begin{equation}
    \text{ACC} = \frac{\sum_{i=1}^{N} (\hat{x}_i - \bar{\hat{x}})))(x_i - \bar{x})}{\sqrt{\sum_{i=1}^{N} (\hat{x}_i - \bar{\hat{x}}}) \cdot \sqrt{\sum_{i=1}^{N} (x_i - \bar{x})^2}},
\end{equation}
where $\hat{x}_i^2$ is the predicted value at pixel $i$ and $x_i$ is the true weather state at pixel $i$. $N$ is the number of pixels in one frame. $\bar{\hat{x}}$ is the mean of forecast values, and $\bar{x}$ is the climatology (i.e., long-term mean of weather states, which is estimated on 39 years ERA5 data). The nRMSE between target and prediction is also computed after centering and scaling all data using their mean and standard deviation. Besides, we provide the CSV files containing the sample-wise RMSE and sample-averaged RMSE.

Quantile errors are defined as the Relative Quantile Error (RQE). A positive RQE value indicates that extreme values are overestimated, while a negative value indicates underestimation \citep{xu2024extremecast}. The perfect model would achieve an RQE of 0. The RQE at a $\alpha$ percentile is computed as follows:
\begin{align*}
\text{RQE}_{\alpha} &= \frac{|Q_{\alpha}(\mathbf{\hat{x}}) - Q_{\alpha}(\mathbf{x})|}{Q_{\alpha}(\mathbf{x})}, \\
Q_{\alpha}(\mathbf{z}) &= \inf \{ z \in \mathbb{R} : F_{\mathbf{z}}(z) \geq \alpha \},
\end{align*}
where $F_{\mathbf{z}}(z)$ is the cumulative distribution function of $\mathbf{z}$. The percentiles we choose are from 80 to 98.

\subsubsection{Tropical Cyclones}
Similar to heatwaves and cold waves, the model performance on tropical cyclones is evaluated by RMSE and ACC. 

\subsubsection{Storms and Extreme Precipitations}

We introduce the widely adopted metrics for precipitation nowcasting, that is, Probability of Detection (POD), False Alarm Ratio (FAR), Critical Success Index (CSI), and Heidke-Skill-Score (HSS) \citep{shi2017deep}. First, the rain map is binarized by thresholding at 0.5, 2, 5, 10, and 30 $mm/h$. Let: $a$ be the number of true positives (hits), $b$ be the number of false positives (false alarms), $c$ be the number of false negatives (misses), $d$ be the number of true negatives (correct rejections), $N=a+b+c+d$ be the total number of samples, then $\text{POD} = \frac{a}{a + c}$,  $\text{FAR} = \frac{b}{a + b}$, $\text{CSI} = \frac{a}{a + b + c}$, $\text{HSS} = \frac{2(ad - bc)}{(a + c)(c + d) + (a + b)(b + d)}$ at different thresholds are computed. 

\subsubsection{Fires}
The binary segmentation performance is evaluated against the F-1 Score and Intersection over Union (IoU). 
\begin{equation}
    \text{F-1} = \frac{2a}{2(a+d) + b + c},
\end{equation}
 and
\begin{equation}
    \text{IoU} = \frac{|\hat{Y} \cap Y|}{|\hat{Y} \cup Y|},
\end{equation}
where $\hat{Y}$ is the predicted set (e.g., pixels predicted as fire scars), and $Y$ is the ground truth set (e.g., pixels that are actually fire scars). We also provide the CSV files storing the sample-wise F-1 score and sample-wise IoU, which can be used to compute the sample-averaged F-1 and sample-averaged IoU. 

\subsubsection{Flood}
The flood detection performance is evaluated by the F-1 score and the IoU for each class (excluding the non-flood area). We also report the weighted F-1 / IoU considering the class imbalance and an unweighted average of F-1 / IoU for both open-flood and urban-flood areas (ignoring the non-flood area). 

\subsection{Appendix: Experiments}\label{app:experiments}
Table~\ref{tab:models} summarizes all the models evaluated in our experiments. In the following experiments, unless specified otherwise, we load the pre-trained weights into the encoder and fine-tune only the decoder to evaluate the effectiveness of pre-trained features. To accommodate data with diverse modalities, we modified the FM's first layer, making every effort to retain as much of the pre-trained weights as possible (e.g., by copying weights to the new layer). The U-Net, however, is randomly initialized and trained end-to-end.

\subsubsection{Heatwaves and cold waves}
The models are trained with MAE loss ($L_1$) between the target and output. All data is normalized using the mean and standard deviation of the dataset. The model input channel is set as 4, including $t_{2m}$ and three masks (topography, land-sea mask, soil type mask). The output channel is set to 1.

We use a constant learning rate of \(5 \times 10^{-3}\) (\(1 \times 10^{-7}\) for cold wave) with a weight decay of \(5 \times 10^{-6}\). The batch size is set to 16, and the input images are resized and cropped to \(224 \times 224\) (\(96 \times 96\) for cold wave). Training is performed using the AdamW optimizer \citep{loshchilov2017fixing}. For heatwave detection, the training runs for 20 epochs. Given the relatively small number of cold wave events, the models are prone to overfitting. To mitigate this, we employ early stopping, selecting the best-performing model on 30\% of the training data with a patience threshold of 5 epochs.

\subsubsection{Tropical Cyclones}
Due to the high-dimensionality of the input (4D and 5D batched input) and varying sizes of inputs, only the Aurora model was tested for this case. The model was trained using the weighted $L_1$ loss. We use the same weights for each variable and pressure levels as in the pretraining stage.

The model is trained for 8,000 steps at a constant learning rate of \(5 \times 10^{-6}\) and a weight decay of \(5 \times 10^{-6}\), optimized using the AdamW optimizer.

\subsubsection{Storms and Extreme Precipitations}
The models are trained using the $L_1$ loss. We train the model using single-step output. The input sequence length is 2 or 3, and the output sequence length is 1. For the model that has no temporal module, we simply take the temporal dimension as the channel dimension. Therefore, their input channel is set as $L_{in}$, and the output channel is set as $L_{out}$. 

Among the baseline models, Aurora and Prithvi have temporal modules to process sequential images. The models are trained for a total of 20,000 steps. We use a (half) cosine decay with a linear warm-up from zero for 800 steps. The base learning rate is \(1 \times 10^{-5}\) (\(1 \times 10^{-5}\) for extreme precipitation), which the schedule reduces by a factor of 100 at the end of training. For extreme precipitation, we evaluate both a fully fine-tuned Aurora and a LoRA fine-tuned Aurora for forecasting horizons of 5, 3, and 0.5 hours. Additionally, Prithvi is trained at the 3- and 0.5-hour forecasting horizons. The batch size is set to 16 for all experiments.  

We also fine-tune U-Net, Segformer, and ConvNeXt by treating the temporal dimension as the channel dimension. However, the model failed to produce useful outputs. Therefore, the experimental settings and results are ignored in this paper.

\subsubsection{Fires}
Considering the imbalanced label distribution, we use Dice loss \citep{sudre2017generalised} to train the model. The input channel is 6, and the output channel is 2.

For SatMAE and Prithvi, we fine-tune the models for 50 epochs. The AdamW optimizer is used with a constant learning rate of \(1.3 \times 10^{-5}\) and a weight decay of \(5 \times 10^{-6}\). For other models, fine-tuning is conducted over 20 epochs, employing an initial learning rate of \(5\times 10^{-5}\) with a cosine learning rate scheduler. For DOFA* and Prithvi-2, we utilize features from layers 5, 11, 17, and 23.

To examine how varying the number and combination of features from different depths of FMs affects the final performance, we use two settings for DOFA. In the first setting, DOFA employs a CNN-based decoder with the deepest feature as input. In the second setting, DOFA* utilizes features from four output layers (7, 11, 15, and 23)), which are then decoded into the final output using UperNet. To analyze the impact of fine-tuning approaches and the value of pre-trained weights, we also compare the model's performance using randomly initialized weights, pre-trained weights with full fine-tuning, and pre-trained weights with fine-tuning applied only to the encoder.

\subsubsection{Flood}
We use Dice loss to train the model. The input channel is 8 and the output channel is 3. For SatMAE, Prithvi, ClimaX, and UNet, we train the models over 50 epochs using a constant learning rate of 2e-5. We select the best-performing model on 15\% of the training samples. For DOFA*, Segformer, Convnxt, and Prithvi2, we train the model for 30 epochs, employing an initial learning rate of \(1 \times 10^{-3}\) with a cosine learning rate scheduler. We repeat the experiments across five random seeds and present the mean and standard deviation of the resulting scores. The rest experimental setup is the same as that used in the Fires task.

\subsection{Results and Discussion}\label{app:results}
We conduct extensive experiments on the benchmark dataset, the details of which are outlined below.
\subsubsection{Heatwaves}
\paragraph{Results} Table~\ref{tab:res_heatwave} is the ACC, nRMSE, and RMSE of different models at a 10-day forecasting horizon. All models achieve an ACC score greater than 0.5. Aurora achieves the best scores with ACC at 0.89 and RMSE at 4.5, which is not surprising given that its pretraining datasets include ERA5 data. DOFA, which achieves the second-best score, was pretrained on five modalities and demonstrates better transferability when fine-tuned on unseen image types.

\begin{figure}[h!]
    \centering
    \includegraphics[width=.5\textwidth]{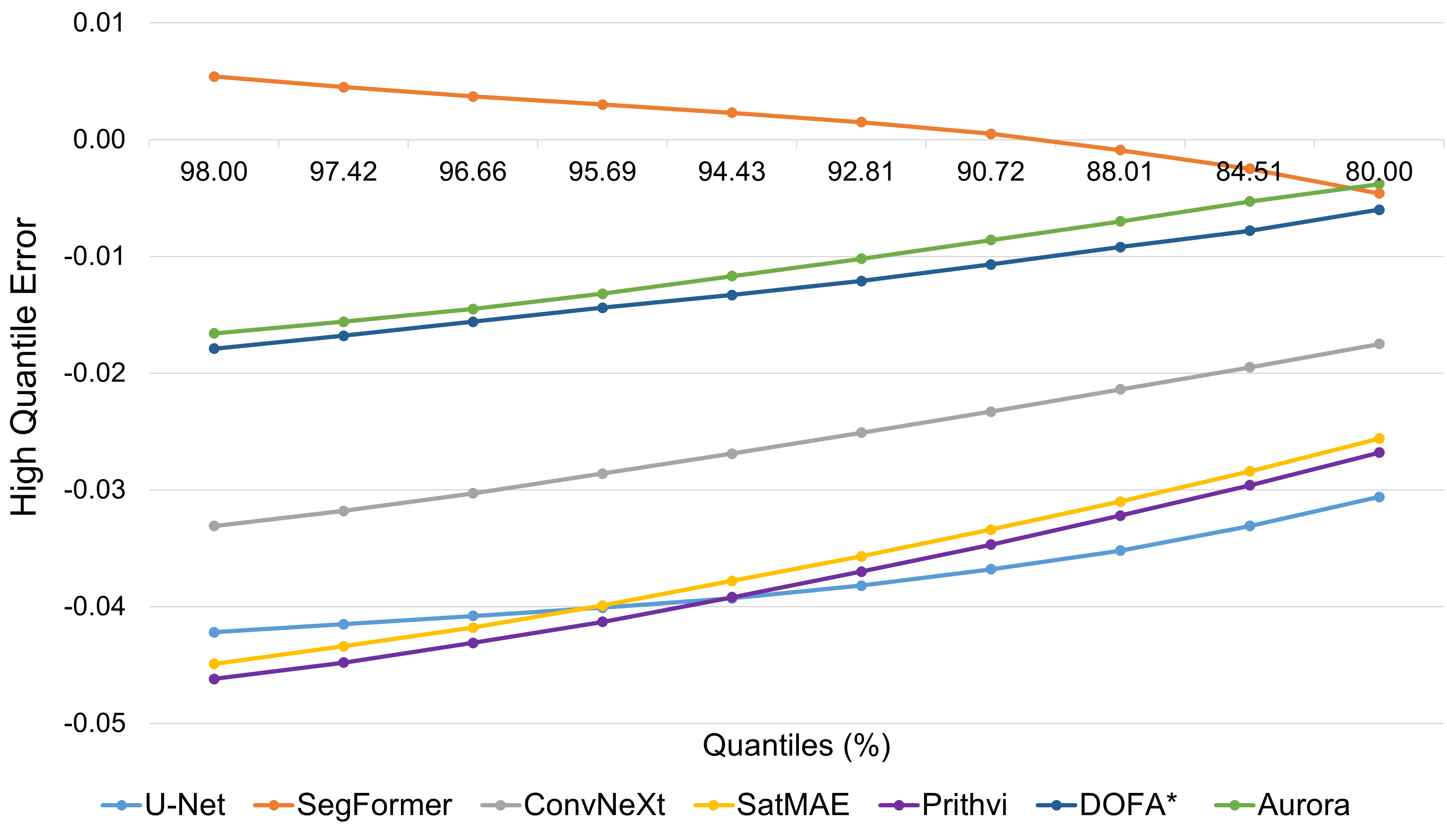}
    \caption{\small Heatwave: 80 to 98 quantile errors of different models for forecasting window of 10 days in advance.}
    \label{fig:hqe_hw}
\end{figure}

As shown in the RQE (80\% to 98\%) in Fig.~\ref{fig:hqe_hw}, Segformer tends to overestimate the high temperatures, while the other models tend to underestimate the temperatures. Aurora, DOFA, and Segformer exhibit smaller RQE, indicating their better ability to capture extreme values.

\begin{figure}[h!]
    \centering
    \includegraphics[width=1\textwidth, trim={0cm 10cm 0cm 0cm},clip]{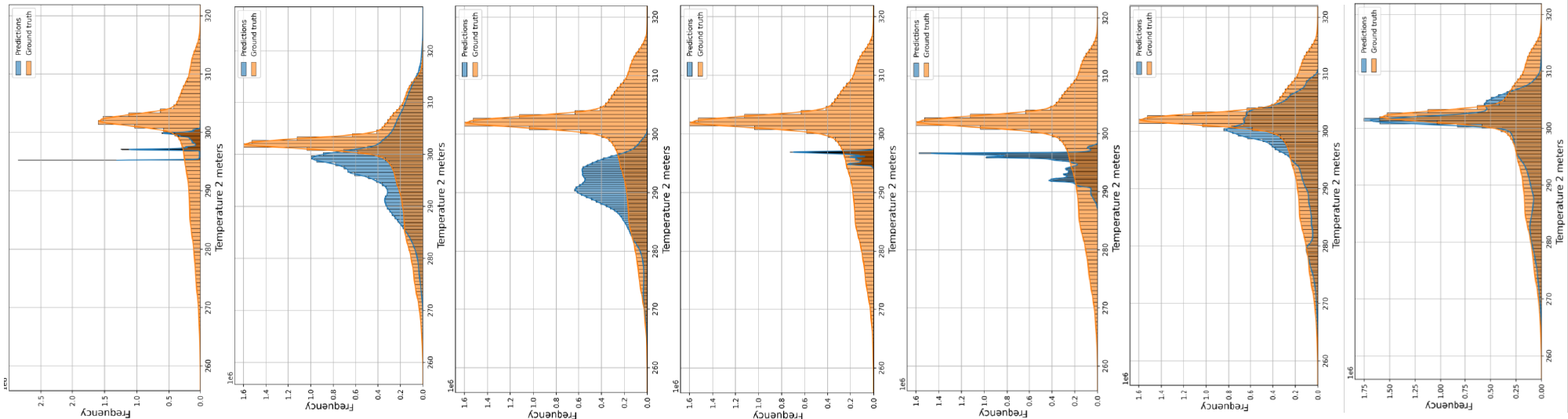}
    \caption{\small Heatwave: Histogram of the labels and predictions of different models for forecasting window of 10 days in advance.}
    \label{fig:hist_hw}
\end{figure}

The histogram shows similar results, with Segformer, DOFA, and Aurora being able to approximate the distribution. In contrast, the other models tend to produce smoothed outputs, as evident in the histogram in Fig.~\ref{fig:hist_hw}, where most values are located within a very narrow range.

\begin{figure}[h!]
    \centering
    \includegraphics[width=1\textwidth, trim={0 5.1cm 10cm 0cm},clip]{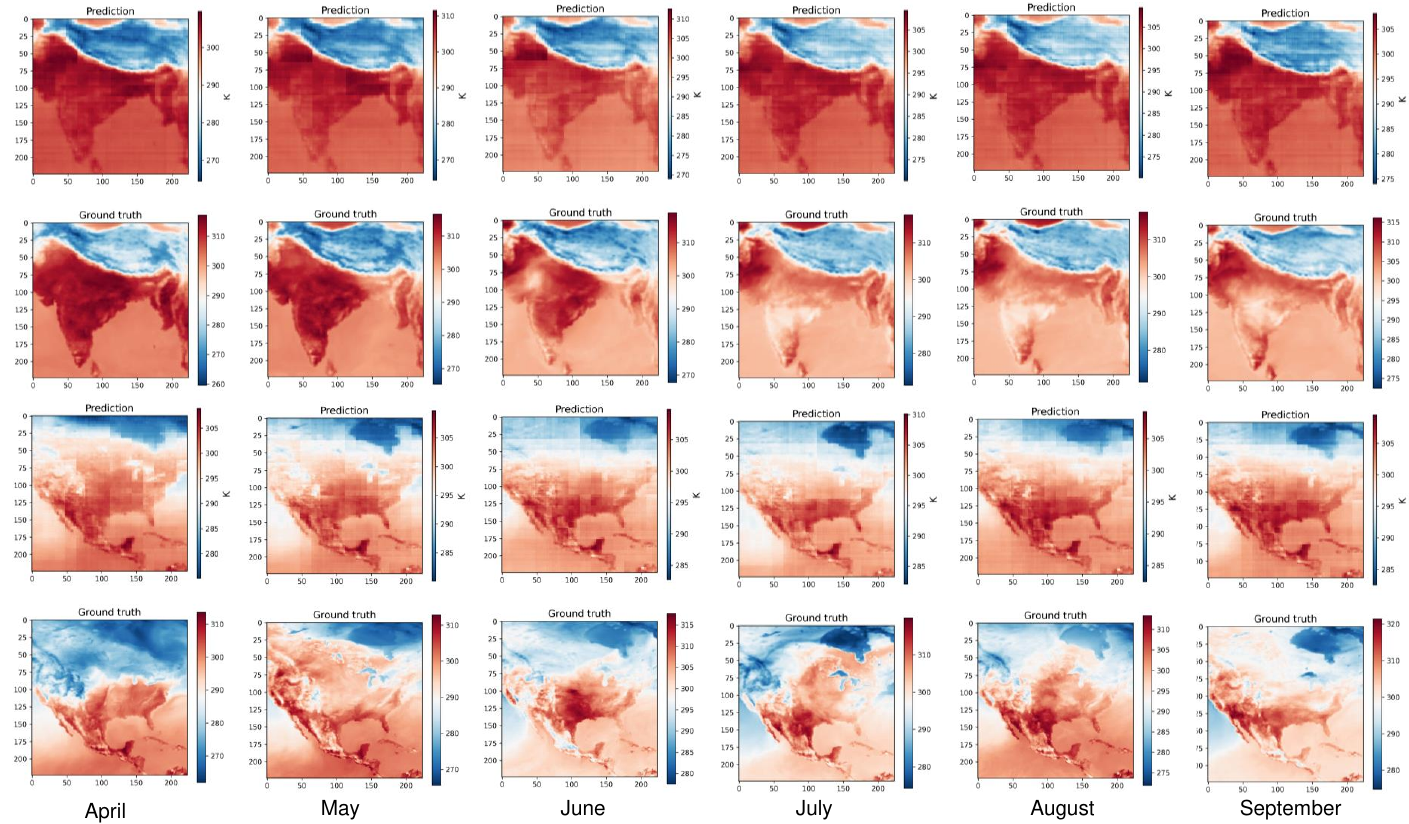}
    \caption{\small Heatwave: Visual sample images of model (Aurora) output for forecasting window of 10 days in advance in April to September.}
    \label{fig:vis_hw}
\end{figure}

Figure~\ref{fig:vis_hw} shows sample frames from the best-performing model. It is noticeable that the model can capture geographic patterns.

\paragraph{Discussion} The experimental results on heatwave events can evaluate the FM's transferability from different pretrained data types (vision, remote sensing, and weather) to the downstream data type (weather). In summary, the generalization capacity of the FM on unseen data modalities highlights the potential for bridging the gap between EO, W\&C, and vision images. Nevertheless, end users should select an FM that has been trained on the data modality relevant to the downstream task. Lastly, incorporating additional data modalities could potentially improve the model's generalizability.

\begin{table}[h!]
    \centering
    \resizebox{0.8\textwidth}{!}{
    \begin{tabular}{l r r r r r r r r}
        \toprule
        Model & U-Net & SegFormer & ConvNeXt & SatMAE & Prithvi & DOFA* & Aurora \\ 
        \midrule
         ACC $\uparrow$ & 0.5916 & 0.7395 & 0.8002 & 0.6208 &  0.5948 & 0.8060 &  \textbf{0.8887}\\ 
        nRMSE $\downarrow$ & 0.9740 & 0.7045 & 0.7137 & 0.9789 & 0.9415 & 0.6277 & \textbf{0.4871}  \\ 
        RMSE $\downarrow$ & 9.0342 & 6.5347 & 6.6200 & 9.0803 & 8.7335 & 5.8220 & \textbf{4.5179}  \\
        \bottomrule
    \end{tabular}
    }
    \caption{\small Heatwave: ACC, nRMSE, and RMSE performance of different models for forecasting window of 10 days in advance.}
    \label{tab:res_heatwave}
\end{table}
\begin{figure}[h!]
    \centering
    \includegraphics[width=1\textwidth, trim={0 11.8cm 2.1cm 0cm},clip]{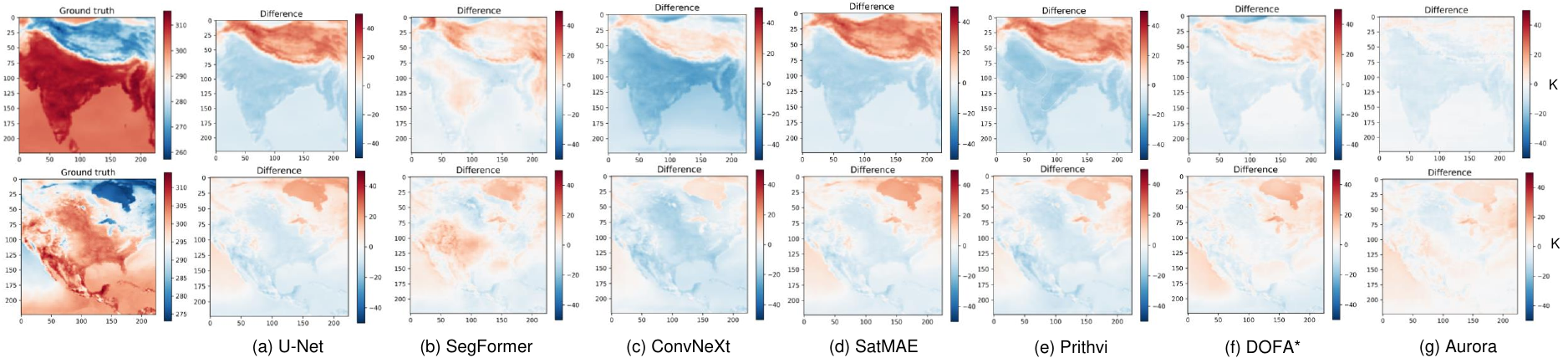}
    \caption{\small Heatwave: spatial bias of model outputs, computed as the difference between predictions and ground truth at a 10-day lead time.}
    \label{fig:bias_hw}
\end{figure}

\subsubsection{Cold waves}
\paragraph{Results} Table~\ref{tab:res_coldwave} is the ACC, nRMSE, and RMSE of different models at a 10-day forecasting horizon. The error scores are generally higher than those observed in the heatwave case. Only Aurora achieves an ACC greater than 0.5, with a score of 0.776. However, U-Net achieved a better error score in the coldwave case compared to the heatwave case. For instance, its nRMSE reduces 0.3363 points. This could be attributed to U-Net's relatively smaller number of parameters, which makes it less prone to overfitting in the smaller cold waves dataset. Most models failed to surpass the baseline (randomly initialized U-Net).

\begin{figure}[h!]
    \centering
    \includegraphics[width=.5\textwidth]{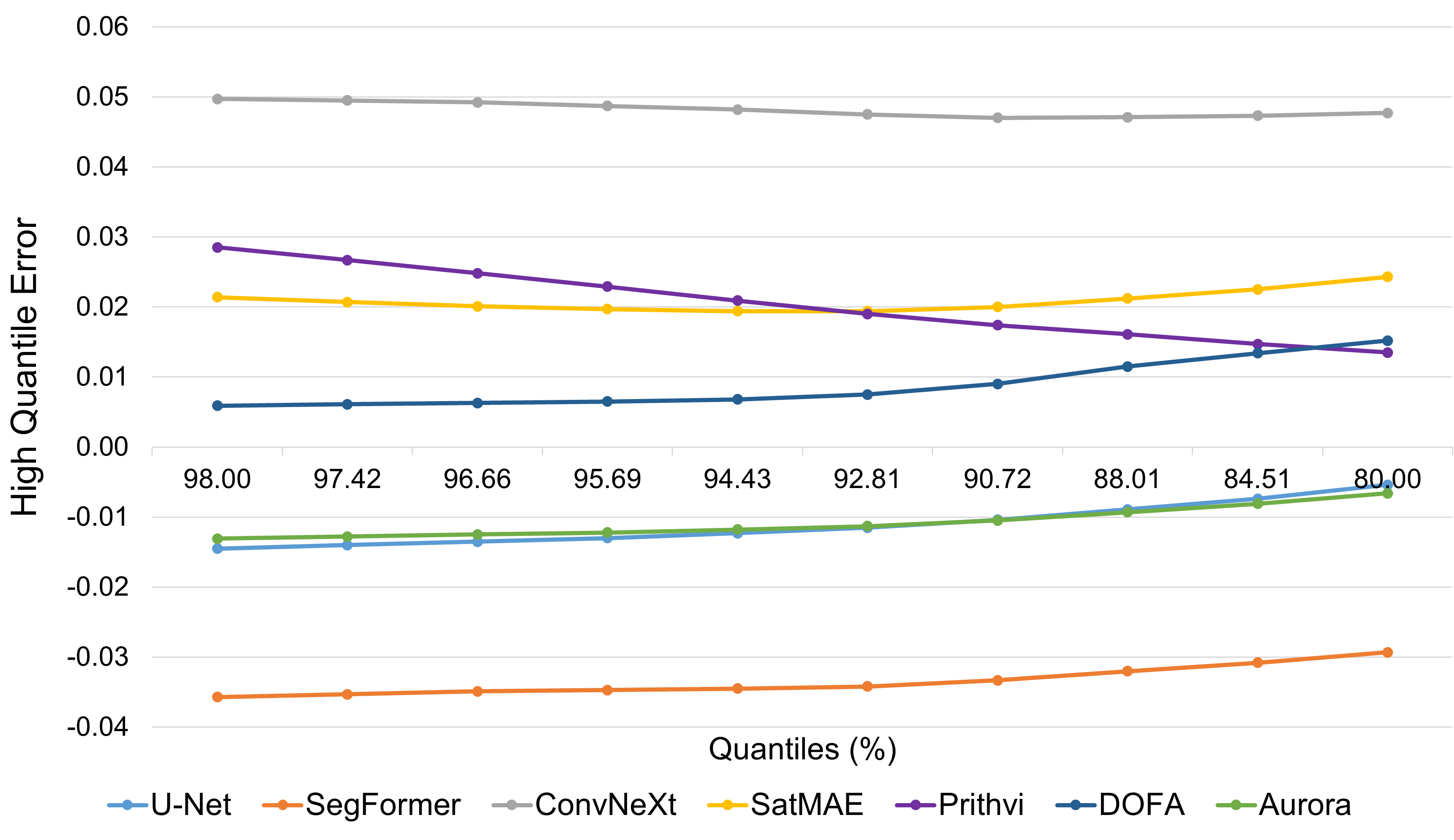}
    \caption{\small Cold waves: 80 to 98 quantile errors of different models for forecasting window of 5 days in advance.}
    \label{fig:hqe_cw}
\end{figure}

Regarding the RQE in Fig.~\ref{fig:hqe_cw}, Aurora, U-Net, and Segformer tend to underestimate high temperatures, while the other models tend to overestimate them.  

\begin{figure}[h!]
    \centering
    \includegraphics[width=1\textwidth, trim={0 9.2cm 0 0cm},clip]{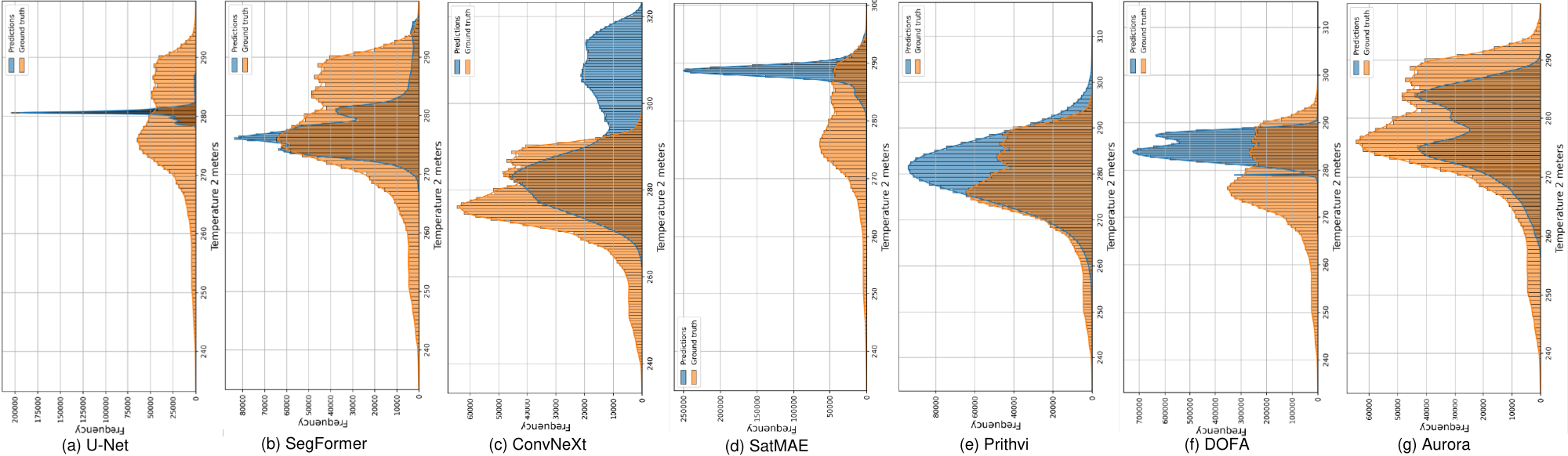}
    \caption{\small Cold waves: Histogram of the labels and predictions of different models for forecasting window of 5 days in advance.}
    \label{fig:hist_cw}
\end{figure}

Only Aurora is able to capture both peaks in the histogram, as shown in Fig.~\ref{fig:hist_cw}. This demonstrates its superior capacity to capture underlying distribution shifts.

\paragraph{Discussion} Compared to heatwaves, the experiments on cold waves primarily focus on evaluating model performance in relation to downstream training data volume. The failure of other models suggests that FMs may struggle when dealing with very limited labeled data, especially with unseen data modalities. As a result, techniques to prevent overfitting should be considered to improve model performance in such scenarios.

\begin{table}[h!]
    \centering
    \resizebox{0.8\textwidth}{!}{
    \begin{tabular}{lrrrrrrrrr}
        \toprule
        Model & U-Net & SegFormer & ConvNeXt & SatMAE & Prithvi & DOFA & Aurora \\
        \midrule
         ACC $\uparrow$ &  0.3983 & 0.2455 & 0.4727 & 0.4013 & 0.2301 & 0.4340 & \textbf{0.7760} \\
        nRMSE $\downarrow$ & 0.6377 & 0.7289& 1.3402 & 0.9421 & 0.8791 & 0.7132 & \textbf{0.3731}\\
        RMSE $\downarrow$ & 8.3730 & 9.5694 & 17.5960 & 12.3689 & 11.5412 & 9.3639 & \textbf{4.8990}\\

        \bottomrule
    \end{tabular}
    }
    \caption{\small Cold waves: ACC, nRMSE, and RMSE performance of different models for forecasting window of 5 days in advance.}
    \label{tab:res_coldwave}
\end{table}

\subsubsection{Tropical Cyclones}
Tropical cyclone prediction is a more challenging task due to the involvement of multiple variables and pressure levels. As a result, we have only evaluated Aurora, which is capable of handling randomly sized inputs.  

\paragraph{Results} The error scores in Table~\ref{tab:res_tc_sur} and \ref{tab:res_tc_atm} at the 6-hour forecasting horizon demonstrate that Aurora can achieve satisfactory performance across all predicted variables, particularly in wind prediction.  

\begin{figure}
    \centering
    \includegraphics[width=1\textwidth, trim={0 8cm 3.5cm 0cm},clip]{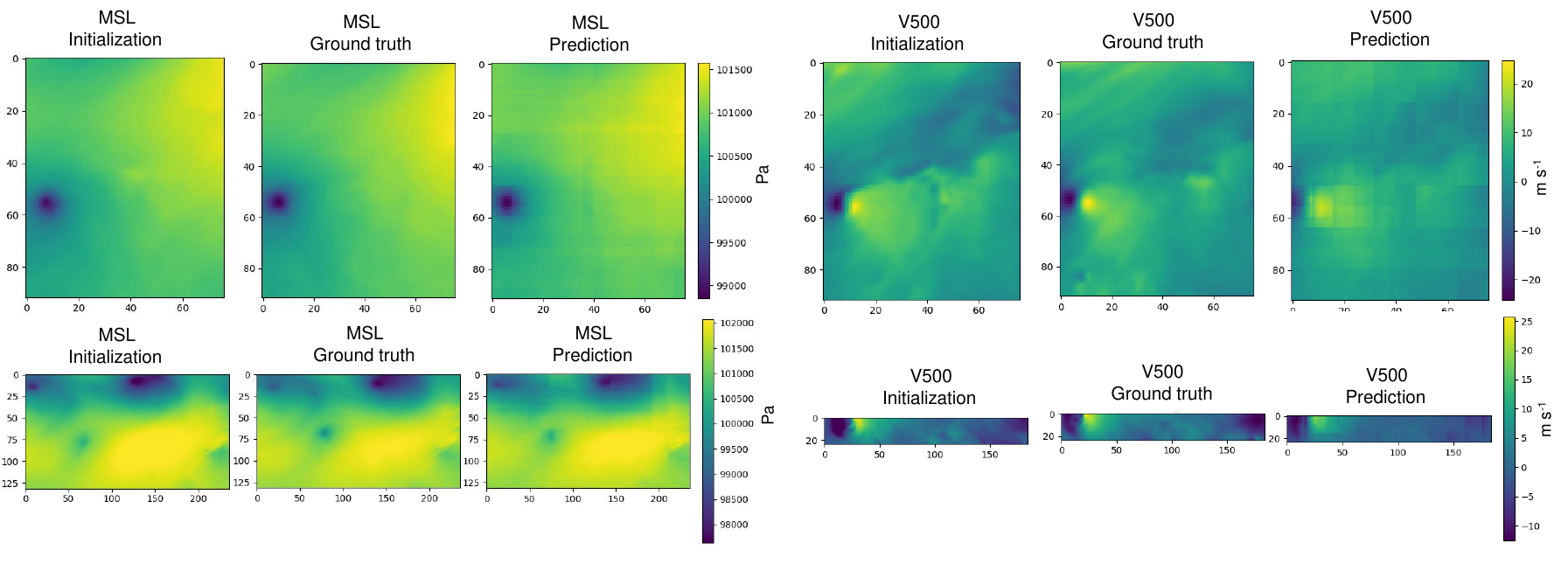}
    \caption{\small Tropical cyclones: The visual sample frames for the forecasting window of 6 hours in advance.}
    \label{fig:vis_tc}
\end{figure}

Figure~\ref{fig:vis_tc} illustrates that the model successfully captures changes in the local minimum or maximum values, which is crucial for accurately tracking the motion of the cyclone's eye.

\paragraph{Discussion} The experiments on tropical cyclones demonstrate the effectiveness of FMs in fine-tuning from global forecasts to local phenomena while handling complex, multivariate weather conditions.

\begin{table}[h!]
    \centering
        \resizebox{0.4\textwidth}{!}{
    \begin{tabular}{l r r r }
        \toprule
         Variable & MSL & U10 & V10 \\ 
         \midrule
        ACC $\uparrow$ & 0.8718 & 0.9094 & 0.8977 \\
        RMSE $\downarrow$ & 183.9857 & 1.9625 & 2.2034 \\
        \bottomrule
         \end{tabular}
          }
    \caption{\small Tropical cyclones: ACC and RMSE performance on surface variables for forecasting windows of 6 hours in advance.}
    \label{tab:res_tc_sur}
\end{table}    

\begin{table}[h!]
\centering
    \resizebox{0.65\textwidth}{!}{
\begin{tabular}{l r r r r r r }
    \toprule
    Metrics & \multicolumn{3}{c}{ACC $\uparrow$} & \multicolumn{3}{c}{RMSE $\downarrow$}\\ \hline
    Variable & Z & U & V & Z & U & V\\ 
    \midrule
    1000 $hPa$ & 0.8614 & 0.9073 & 0.8985 & 151.7871 & 2.2763 & 2.5259 \\ 
    850 $hPa$ & 0.7786 & 0.9212 & 0.8859 & 145.0797 & 2.6457 & 2.7557 \\
    700 $hPa$ & 0.8250 & 0.9315 & 0.8666 & 145.5359 & 2.6247 & 2.6522 \\
    500 $hPa$ & 0.9738 & 0.9497 & 0.8675 & 146.3311 & 2.7534 & 2.9393 \\
    200 $hPa$ & 0.9956 & 0.9579 & 0.8329 & 186.8965 & 4.6705 & 4.0715 \\
    \bottomrule
\end{tabular}}
\caption{\small Tropical cyclones: ACC and RMSE performance on atmospheric variables for forecasting windows of 6 hours in advance.}
\label{tab:res_tc_atm}
\end{table}

\subsubsection{Storms}
In the storm nowcasting task, we compare the temporal processing capabilities of Prithvi and Aurora.

\paragraph{Results} In Table~\ref{tab:res_storm_probscore}, Prithvi demonstrates better FAR than Aurora across all thresholds. However, for other metrics, Aurora outperforms Prithvi at thresholds of 0.5, 2, 10, and 30 $mm/h$. At the 5 $mm/h$ threshold, Prithvi outperforms Aurora across all skill scores.

\begin{figure}[h!]
    \centering
    \includegraphics[width=1\textwidth, trim={0 0cm 6.4cm 0cm},clip]{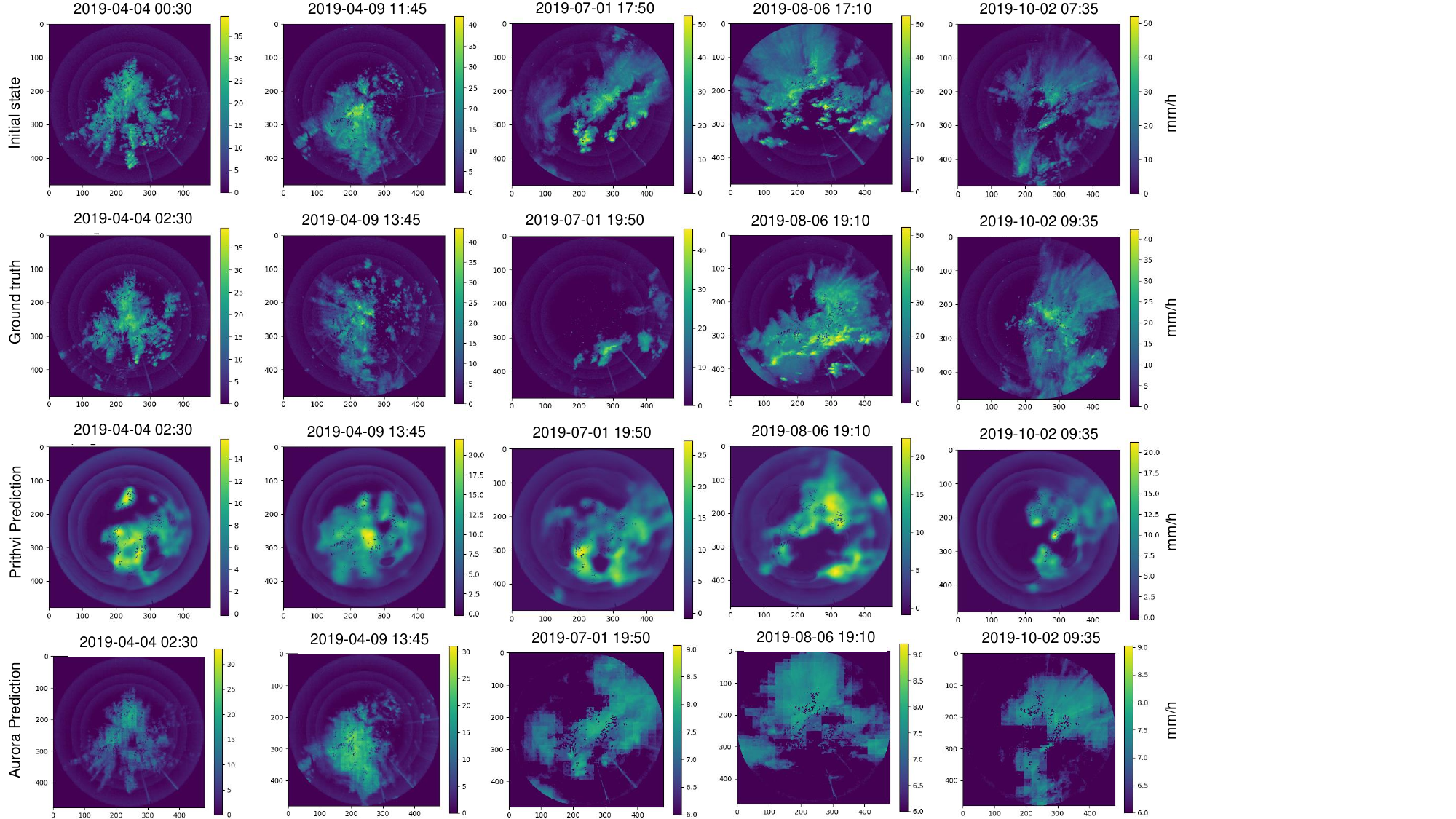}
    \caption{\small Storms: The visual comparison of sample frames between different models (Prithvi and Aurora) for a nowcasting window of 2 hours in advance.}
    \label{fig:vis_storm}
\end{figure}

Figure~\ref{fig:vis_storm} presents sample visual frames at a 2-hour horizon. It is observed that Prithvi excels at capturing temporal dynamics, while Aurora performs better in generating fine-resolution outputs, albeit with a tendency to reproduce the last frame of the input. Both models tend to underestimate extreme precipitation values.

\paragraph{Discussion} The experimental results on storms assess the transferability of FMs across the spatial domain and unseen variables. The results show that some FM can overcome large gaps between the pre-trained data and downstream data, such as those from 0.25$^\circ$ coarse resolution data to 500 meters of fine spatial granularity, from global coverage to regional phenomena, and extend to new, unseen variables. However, there is still room for improvement in capturing spatial-temporal dynamics, as well as addressing the underestimation of extreme values.

\begin{table}[h!]
    \centering
    \resizebox{0.8\textwidth}{!}{
    \begin{tabular}{l r r r r r r  }
        \toprule
        Model & Thresholds &  0.5 $mm/h$ & 2 $mm/h$ & 5 $mm/h$ & 10 $mm/h$ & 30 $mm/h$ \\ 
        \midrule
        \multirow{4}{*}{Prithvi (temporal)} & POD $\uparrow$ &  0.0828 & 0.0128 & \textbf{0.0012} & 0.0002 & 0 \\
        ~ & FAR $\downarrow$ & \textbf{0.5009} & \textbf{0.6837}  & \textbf{0.7838} & \textbf{0.4197} & nan\\
        ~ & CSI $\uparrow$ & 0.0764 & 0.0124 & \textbf{0.0012} & 0.0002 & 0 \\
        ~ & HSS $\uparrow$ & 0.1303 & 0.0234 & \textbf{0.0023} & 0.0003 & 0\\ \hdashline
        \multirow{4}{*}{Aurora} & POD $\uparrow$ & \textbf{0.1536} &  \textbf{0.0145} & 0.0007 & \textbf{0.0004} &  \textbf{0.0001}  \\
        ~ & FAR $\downarrow$ & 0.6528  & 0.7882  &  0.8215 &  0.8330 &  \textbf{0.9463} \\
        ~ & CSI $\uparrow$ & \textbf{0.1192}  & \textbf{0.0138} & 0.0007 & \textbf{0.0004} & \textbf{0.0001} \\
        ~ & HSS $\uparrow$ & \textbf{0.1895}  & \textbf{0.0253} & 0.0014 & \textbf{0.0007}  & \textbf{0.0003} \\ 
        \bottomrule
    \end{tabular}}
    \caption{\small Storms: probability scores of the different models for a nowcasting window of 2 hours in advance.}
    \label{tab:res_storm_probscore}
\end{table}

\subsubsection{Extreme Precipitations}
The extreme precipitation dataset describes the same climate variable (precipitation) as the storm dataset, albeit at significantly different spatial resolutions and coverage.

\paragraph{Results} 
Comparing Table~\ref{tab:res_expcp_probscore} to Table~\ref{tab:res_storm_probscore}, Prithvi has better POD and CSI compared to the storm case. However, it fails to capture the precipitation pattern. Aurora produces both good error scores and visually appealing outputs.

As the forecasting horizon extends, the error scores worsen. We also compared the model performance using different fine-tuning strategies. Compared with fully fine-tuning, LoRA reduces memory consumption and the number of trainable parameters, but it returns slightly worse error scores.

\begin{figure}
    \centering
    \includegraphics[width=1\textwidth, trim={0 4cm 0.5cm 0},clip]{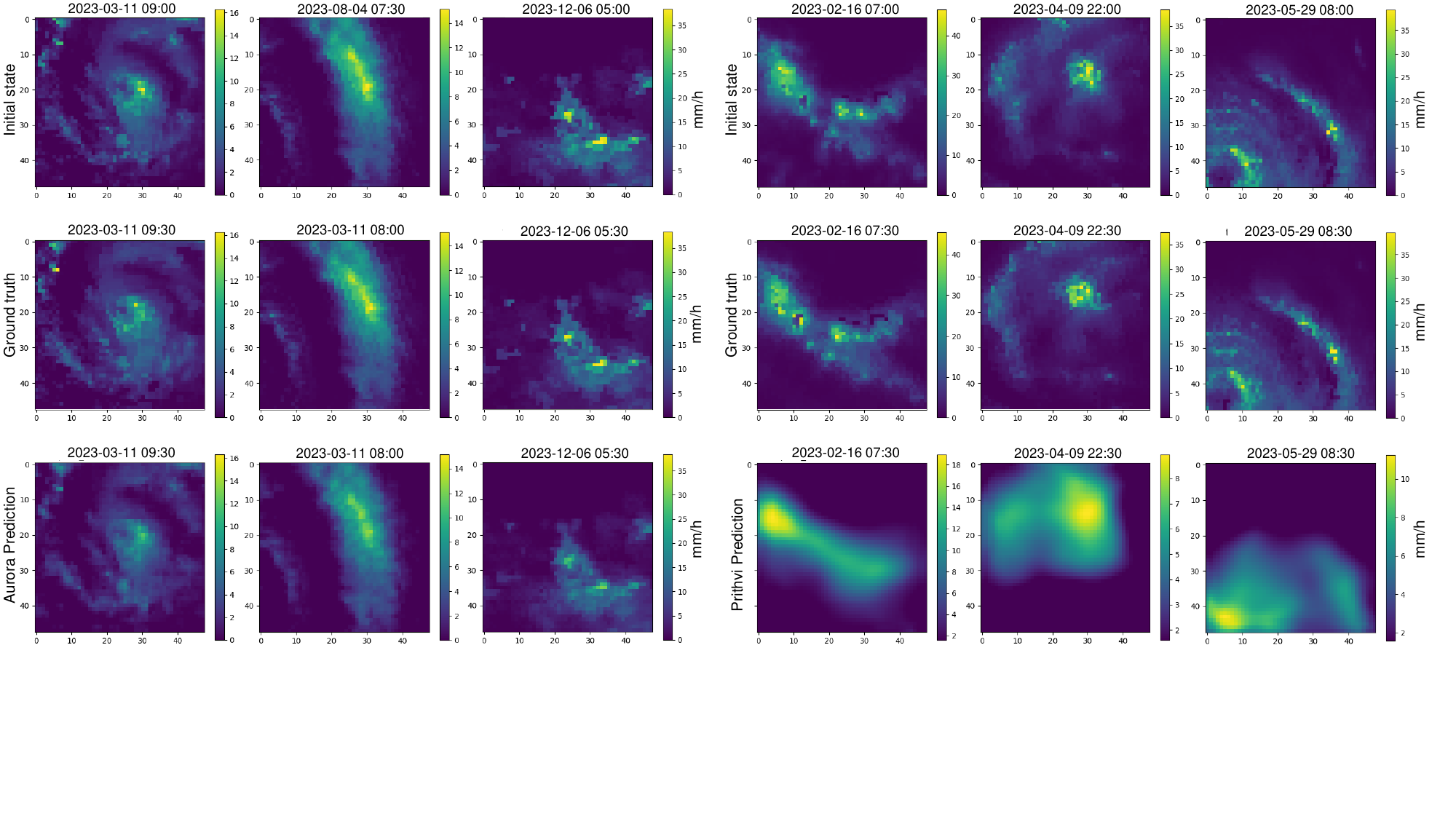}
    \caption{\small Extreme precipitation: The visual comparison of sample frames between Aurora and Prithvi for the nowcasting window of 30 minutes in advance. }
    \label{fig:vis_expcp}
\end{figure}

Figure~\ref{fig:vis_expcp} shows that in a very short forecasting horizon, Aurora generates more consistent spatial patterns with the input data, while Prithvi produces blurred outputs.

\paragraph{Discussion} In contrast to storms, the experimental results on extreme precipitation evaluate the model's transferability to coarse-resolution global forecasts. Besides, it presents challenges in extending forecasting horizons. The experiments also compare parameter-efficient fine-tuning techniques with fully fine-tuned models. The failure of Prithvi in this case indicates that the model struggles with the large gap in spatial granularity (from 20 meters to thousands of kilometers, and from regional to global scales). It could also be due to the interpolation layer used in the decoder. Aurora, which is pre-trained on 0.25$^{\circ}$ data, transfers more easily to 0.1$^{\circ}$ data. Certainly, the performance degrades with an extended forecasting horizon. While different fine-tuning strategies can impact the final result, LoRA still provides a good compromise under limited computational resources. 

\begin{table}[h!]
    \centering
    \resizebox{0.8\textwidth}{!}{
    \begin{tabular}{l r r r r r r  }
        \toprule
        Model & Thresholds &  0.5 $mm/h$ & 2 $mm/h$ & 5 $mm/h$ & 10 $mm/h$ & 30 $mm/h$\\ \midrule  
        \multirow{4}{*}{Prithvi (temporal) (0.5 h)} & POD $\uparrow$ & \cellcolor{orange!10} 1.0 & \cellcolor{orange!30}0.6097 & \cellcolor{orange!30}0.4276 & \cellcolor{orange!30}0.2303 & \cellcolor{orange!90}0 \\
        ~ & FAR $\downarrow$ &  \cellcolor{orange!90}0.5728 & \cellcolor{orange!20}0.2268 & \cellcolor{orange!20}0.2561 & \cellcolor{orange!20}0.2784 & nan \\
        ~ & CSI $\uparrow$ & \cellcolor{orange!90}0.4272 & \cellcolor{orange!30}0.5172 & \cellcolor{orange!30} 0.3727 & \cellcolor{orange!30}0.2115 & \cellcolor{orange!90}0 \\
        ~ & HSS $\uparrow$ & \cellcolor{orange!90}0 & \cellcolor{orange!30}0.6110 & \cellcolor{orange!30}0.5135 & \cellcolor{orange!30}0.3340 & \cellcolor{orange!90}0 \\     
        \multirow{4}{*}{Prithvi (temporal) (3 h)} & POD $\uparrow$ & \cellcolor{orange!10}1.0 & \cellcolor{orange!70}0.3552 & \cellcolor{orange!90}0.0303 & \cellcolor{orange!90}0.0001 & \cellcolor{orange!90}0   \\
        ~ & FAR $\downarrow$ &  \cellcolor{orange!70}0.5698  & \cellcolor{orange!90}0.7650  &\cellcolor{orange!90} 0.9162  & \cellcolor{orange!90}0.9691  & nan \\
        ~ & CSI $\uparrow$ & \cellcolor{orange}0.4304 & \cellcolor{orange!90}0.1647 & \cellcolor{orange!90}0.0227 & \cellcolor{orange!90}0.0001 & \cellcolor{orange!90}0 \\
        ~ & HSS $\uparrow$ & \cellcolor{orange!90} 0 &  \cellcolor{orange!90} 0.0427 &  \cellcolor{orange!90} -0.0002 &  \cellcolor{orange!90} 0 & \cellcolor{orange!90} 0  \\ \hdashline 

        \multirow{4}{*}{Aurora (0.5 h, fully finetune)} & POD $\uparrow$& \cellcolor{orange!30} 0.8423 & \cellcolor{orange!10}0.8199 & \cellcolor{orange!10}0.7430 & \cellcolor{orange!10}0.6546 & \cellcolor{orange!10}0.3170 \\ 
        ~ & FAR $\downarrow$ & \cellcolor{orange!10}0.0943 & \cellcolor{orange!10}0.1523 & \cellcolor{orange!10}0.1828 & \cellcolor{orange!10}0.2149 & \cellcolor{orange!20}0.2384 \\
        ~ & CSI $\uparrow$ & \cellcolor{orange!10}0.7744 & \cellcolor{orange!10}0.7146 & \cellcolor{orange!10}0.6371 & \cellcolor{orange!10}0.5551 & \cellcolor{orange!10}0.2884 \\
        ~ & HSS $\uparrow$ & \cellcolor{orange!10}0.7839 &  \cellcolor{orange!10}0.7912 &  \cellcolor{orange!10}0.7593 &  \cellcolor{orange!10}0.7062 & \cellcolor{orange!10}0.4469 \\
        \multirow{4}{*}{Aurora (0.5 h, LoRA)} & POD $\uparrow$&  \cellcolor{orange!50}0.8292  & \cellcolor{orange!20}0.7604 & \cellcolor{orange!20}0.6170 & \cellcolor{orange!20}0.4052 & \cellcolor{orange!20}0.0465  \\ 
        ~ & FAR $\downarrow$ & \cellcolor{orange!70}0.3633 & \cellcolor{orange!50}0.2902 & \cellcolor{orange!30}0.2997 &  \cellcolor{orange!30}0.2791 & \cellcolor{orange!30}0.2612 \\
        ~ & CSI $\uparrow$ & \cellcolor{orange!30}0.5629 & \cellcolor{orange!20}0.5801 & \cellcolor{orange!20} 0.4881 &  \cellcolor{orange!20}0.3502 & \cellcolor{orange!20} 0.0457 \\
        ~ & HSS $\uparrow$ &\cellcolor{orange!30} 0.4589 &  \cellcolor{orange!20}0.6620  & \cellcolor{orange!20}0.6270  & \cellcolor{orange!20}0.5086 &  \cellcolor{orange!20}0.0872 \\ 
         \multirow{4}{*}{Aurora (3 h, LoRA)} & POD $\uparrow$ &  \cellcolor{orange!70}0.7333  &\cellcolor{orange!50} 0.4379 & \cellcolor{orange!50}0.2038 & \cellcolor{orange!50}0.0840 & \cellcolor{orange!30}0.0001 \\
        ~ & FAR $\downarrow$ & \cellcolor{orange!20}0.2633 & \cellcolor{orange!30}0.3292 & \cellcolor{orange!50}0.3766 &  \cellcolor{orange!70}0.4048  & \cellcolor{orange!10}0 \\
        ~ & CSI $\uparrow$ & \cellcolor{orange!20}0.5811 & \cellcolor{orange!50}0.3604 & \cellcolor{orange!50}0.1815 & \cellcolor{orange!50}0.0794 & \cellcolor{orange!30}0.0001 \\
        ~ & HSS $\uparrow$ & \cellcolor{orange!20}0.5347 & \cellcolor{orange!50}0.4368 &  \cellcolor{orange!50}0.2771 &   \cellcolor{orange!50}0.1409 & \cellcolor{orange!30}0.0003 \\ 
         \multirow{4}{*}{Aurora (5 h, LoRA)} & POD $\uparrow$ & \cellcolor{orange!90}0.6584 & \cellcolor{orange!90}0.2202 & \cellcolor{orange!70}0.0462 & \cellcolor{orange!70}0.0023 &  \cellcolor{orange!90}0 \\
        ~ & FAR $\downarrow$ & \cellcolor{orange!50}0.3290 & \cellcolor{orange!70}0.4003 &  \cellcolor{orange!70}0.4256 &  \cellcolor{orange!50}0.3267 & nan \\
        ~ & CSI $\uparrow$ & \cellcolor{orange!50}0.4978 & \cellcolor{orange!70}0.1920 & \cellcolor{orange!70}0.0446 & \cellcolor{orange!70}0.0023 & \cellcolor{orange!90}0  \\
        ~ & HSS $\uparrow$ & \cellcolor{orange!50}0.4118 &  \cellcolor{orange!70}0.2357 & \cellcolor{orange!70}0.0737 & \cellcolor{orange!70}0.0044 & \cellcolor{orange!90}0   \\ \bottomrule 
    \end{tabular}}
    \caption{\small Extreme precipitation: probability scores of the different models for nowcasting windows of 0.5, 3, and 5 hours in advance. We rank the scores of six models using the color depth of the cell. The deeper the color, the worse the performance of the model.}
    \label{tab:res_expcp_probscore}
\end{table}

\subsubsection{Fires}
\paragraph{Results} In Table~\ref{tab:fire_frozenbody}, when freezing the encoder, Prithvi-2 shows the best performance, as the pre-trained data shares the same source as the downstream data. Models pretrained on ImageNet also deliver promising results. Prithvi-2 outperforms the previous Prithvi model by around 10\%. The DOFA* model outperforms DOFA with an increase of 9.71 points in F1 score and 6.86 points in IoU. Models pretrained on climate data (ClimaX) exhibit the worst performance. Generally, the fully fine-tuned results outperform those obtained with a frozen encoder and randomly initialized weights. When the weights are randomly initialized, the FMs show small variations in performance across different types.

\begin{figure}[h!]
    \centering
    \includegraphics[width=1\textwidth, trim={0 6.4cm 0 0cm},clip]{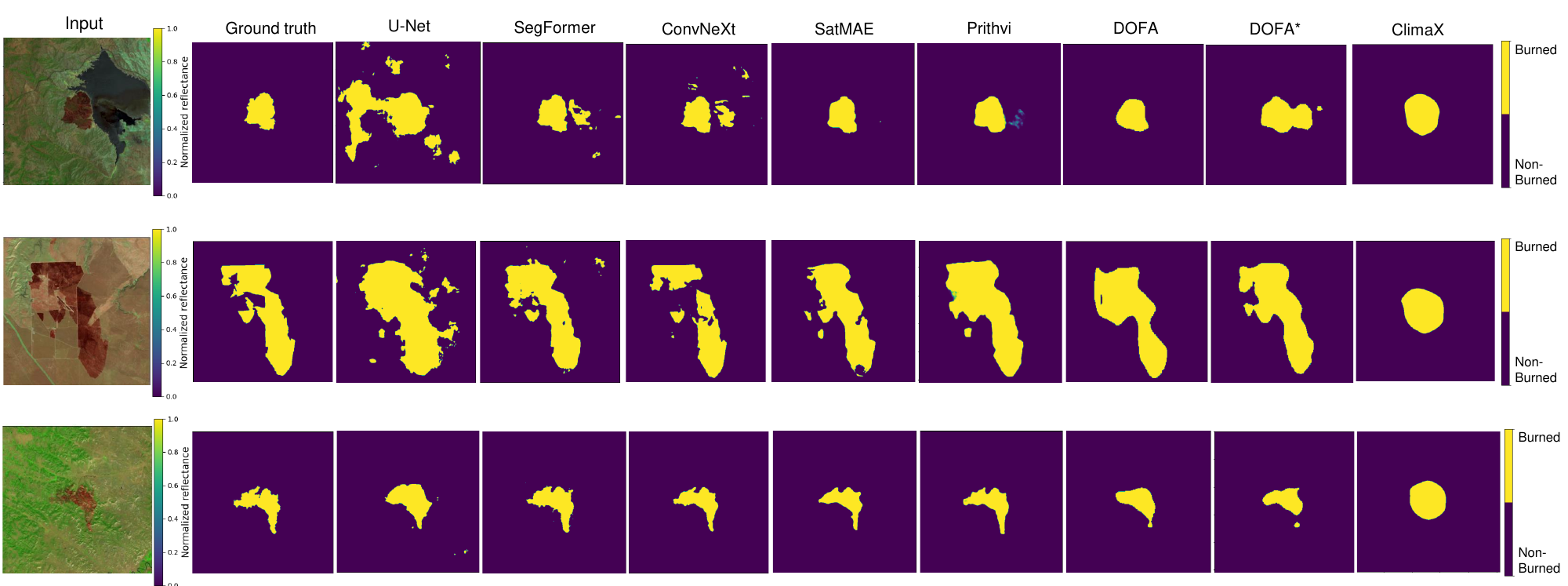}
    \caption{\small Fire prediction maps. We visualize the input using SWIR2 (B12), NIR (B8A), and Red (B04) bands. The yellow pixels in the masks are burned areas.}
    \label{fig:vis_fire}
\end{figure}

Figure~\ref{fig:vis_fire} compares the burned scars maps produced by different models with frozen encoder. U-Net tends to over-segment the image. SegFormer and UperNet, despite providing relatively good outlines, contain some noisy pixels. SatMAE, DOFA, and Prithvi-2 show better segmentation results. ClimaX struggles to learn the spectral information and produces a uniform mask.

\begin{figure}
    \centering
    \includegraphics[width=.6\textwidth]{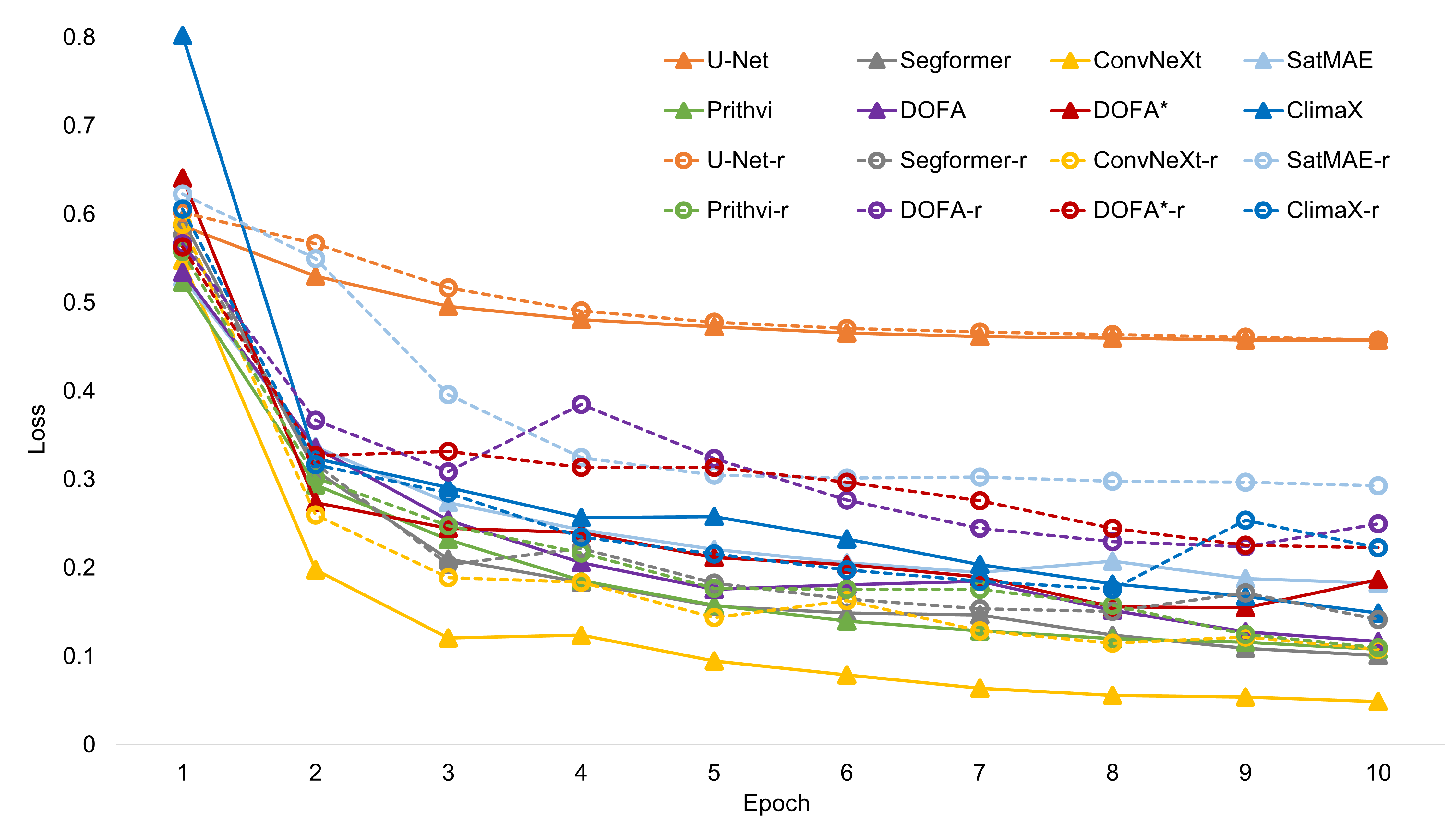}
    \caption{\small Fire: Training loss over the first 10 epochs for different models, fine-tuned with either random initialization (dashed line and marked with ``-r'') or pretrained weights. To ensure comparability, the same learning rate was used across all models. Models with pretrained weights demonstrate faster convergence and achieve lower loss values. The best scores are shown in bold, and the second-best scores are underlined.}
    \label{fig:fire_trainloss}
\end{figure}

Figure~\ref{fig:fire_trainloss} plots the loss curve for the first 10 training epochs. All Models with pretrained weights exhibit faster convergence and attain lower error scores, compared with those with randomly initialized weights.

\paragraph{Discussion} Besides the generalizability of FM to downstream tasks, the experimental design for fires additionally evaluates (1) the value of pretrained weights, (2)the impact of different fine-tuning strategies, and (3) the role of geospatial considerations in model design. FMs pretrained on vision images show promising results when fine-tuned to remote sensing (RS) images, likely because vision images share certain similarities with RS images. However, these vision FMs tend to capture spectral information that is not relevant to the targets. The model pre-trained on weather data failed to generalize to RS images due to the significant disparity between the two data types. In summary, pretrained weights are valuable for faster convergence and better model performance, especially when the downstream task shares the same data modality as the pretraining data. Besides, the geospatial considerations during model design are crucial for geospatial tasks. For example, Prithvi-2, which incorporates the temporal and geolocation embeddings, outperforms Prithvi, underscoring their utility for geospatial applications. Further, among different fine-tuning strategies, the fully fine-tuned model achieves the best performance but at the highest computational cost. With limited training resources, using pretrained weights with a lightweight decoder proves sufficient for producing satisfactory results.


\begin{table}[h!]
    \centering
    \resizebox{1\textwidth}{!}{
    \begin{tabular}{lrrrrrrrr}
        \toprule
        Model & \multicolumn{4}{c}{F1 $\uparrow$} & \multicolumn{4}{c}{IoU $\uparrow$}\\
        Type & Frozen body & Fully finetune & Randomly init. & $\Delta$F1(\%) & Frozen body & Fully finetune & Randomly init. & $\Delta$IoU(\%) 
        \\
        \midrule
        U-Net& - & 83.12 &81.46& 2.04 & - & 83.88& 82.45& 1.73\\
        SegFormer& 69.39& \underline{89.57}& 82.61& 8.41 & 74.09& \underline{89.52} &83.44& 7.29\\
        ConvNeXt& 71.10& 86.19& 86.93& -0.85 & 75.24& \underline{86.61}& 87.08& -0.54\\
        SatMAE& 66.19& 76.59& 48.64& 57.49 & 72.23& 78.89 &62.02& 27.23\\
        Prithvi& 76.45& 78.18& 77.64& 0.70 & 78.82& 80.19& 79.80& 0.49\\
        Prithvi-2& \textbf{82.45}& \textbf{89.95}& \textbf{87.85}& 2.39 &\textbf{ 83.66}& \textbf{89.88}& \textbf{87.99}& 2.15\\
        DOFA &69.25& 83.72& 70.72& 18.37  & 73.80& 84.51 &74.54& 13.37\\
        DOFA* &\underline{78.96}& 87.39 &71.82& 21.65 & \underline{80.66} & 87.51& 88.71& -1.35\\
        ClimaX &44.45 &74.01& 80.32& -7.86 & 59.99 &76.80 &81.62& -5.90\\    
        \bottomrule
    \end{tabular}}
    \caption{\small Fire: F1 (burned scars) and IoU (burned scars) performance of models using different training strategies. The F1 score and IoU are calculated pixel-wise over the test set. $\Delta$F1 and $\Delta$IoU denote the relative change from random initialization to pretrained weights for each model.}
    \label{tab:fire_frozenbody}
\end{table}

\subsubsection{Flood}
\paragraph{Results} In the flood prediction task, the issue of data imbalance is more severe. In Table~\ref{tab:flood_frozenbody}, DOFA* shows the best scores when class imbalance is taken into consideration.

\begin{figure}[h!]
    \centering
    \includegraphics[width=1\textwidth, trim={0 7cm 0 0cm},clip]{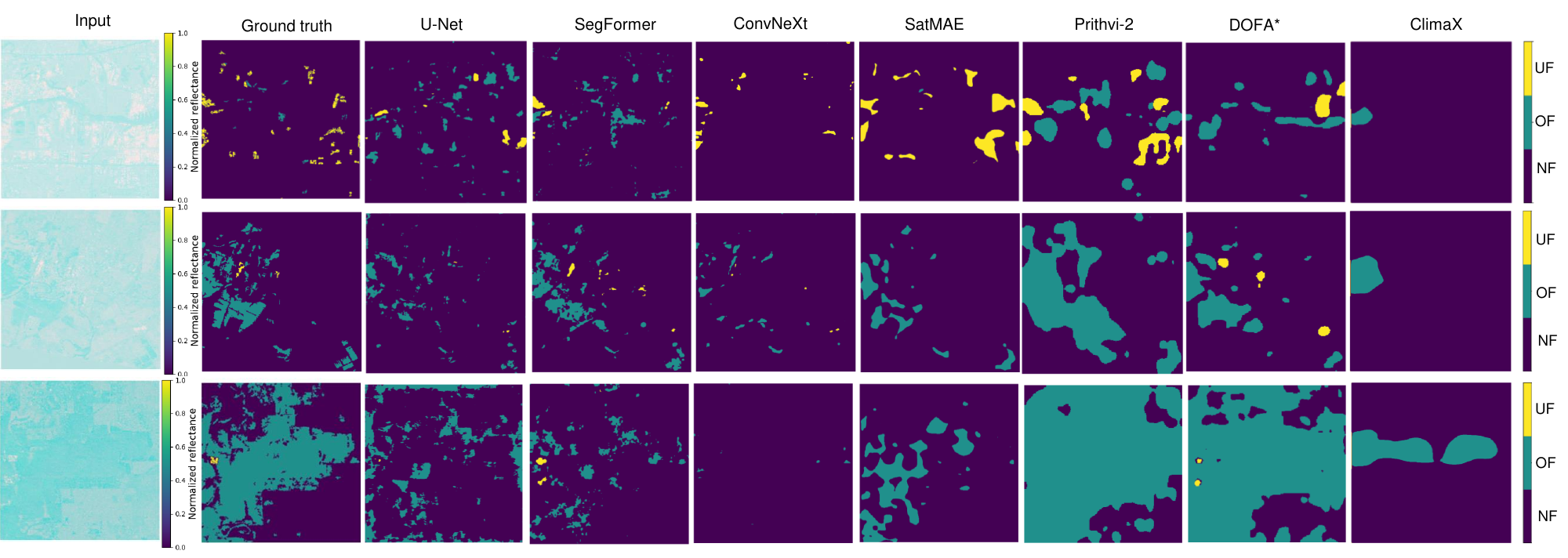}
    \caption{\small Flood Mapping: Normalized input, ground truth, and prediction maps of different models. The input is visualized using VH Sentinel-1 intensity acquired co-event pre, post, and co-event. (UF: urban-flooded, OF: open-flooded, NF: non-flooded)}
    \label{fig:vis_flood}
\end{figure}

Figure~\ref{fig:vis_flood} compares the flood maps generated by different models. Generally speaking, the models' performance is not satisfactory. On the one hand, this could be because the pretrained dataset does not include Sentinel-1 data. Even when pretraining is conducted on Sentinel-1 data, as with DOFA, the model only encodes wavelength information. However, it doesn't consider other sensing features, such as polarization modes and coherence. On the other hand, the temporal information is implicitly embedded within different channels of the data, which complicates the task. Rather than a simple segmentation task, it requires change detection capabilities to differentiate between flooded areas and permanent water bodies. Models that struggle to interpret temporal information (pre-event, co-event, and post-event changes) are prone to errors. For instance, DOFA* misclassified the permanent river as a flooded area, as shown in the first row of Fig.~\ref{fig:vis_flood}. 

\paragraph{Discussion} The experimental results on floods highlight the challenges of label imbalance and and the lack of encoding sensor-specific features, such as coherence in Synthetic Aperture Radar data. They also suggest the need for enhanced temporal processing capabilities in FMs.
\begin{table}[h!]
    \centering
        \resizebox{\textwidth}{!}{
    \begin{tabular}{p{2cm}lp{1cm}p{1cm}p{1cm}p{1cm}p{1cm}p{1cm}p{1cm}p{1cm}p{1cm}p{1cm}}
        \toprule
      
    \multirow{2}{*}{Type} &  \multirow{2}{*}{Model} &  \multicolumn{3}{c}{F1} & \multicolumn{2}{c}{mF1} & \multicolumn{3}{c}{IoU} & \multicolumn{2}{c}{mIoU}\\
      ~ & ~ & Non-flood &  Open flood & Urban flood & Weighted & Macro & Non-flood &  Open flood & Urban flood & Weighted & Macro \\
        \midrule
 \multirow{7}{*}{Frozen body} 
    & SegFormer & 98.02 & 27.03 & 7.94 & 43.90 & 19.87 & 96.12 & 15.63 & 4.14 & 28.83 & 18.97
\\
    & ConvNeXt & \textbf{98.83} &\textbf{38.81} & 6.39 & 43.00 & 18.36 &\textbf{97.69} & \textbf{24.09} & 3.30 & 28.20 & 17.63
\\
    & SatMAE & 94.54 & 10.14& 0.83 & 36.17 & 14.80 & 89.66 & 5.34 & 0.42 & 23.07 & 13.96
 \\
   &  Prithvi & 90.19 & 13.88 & 0.85 & 63.69 & 25.29 & 82.25 & 7.47 & 0.43 & 49.98 & 29.29
 \\
   &  Prithvi-2 & 97.23 & 32.48 & \textbf{8.27} & 65.40 & 28.00 & 94.62 & 19.48 & \textbf{4.34} & 51.19 & 31.70
\\
    & DOFA* & 94.82 & 24.29 & 2.32& \textbf{71.33} & \textbf{30.01} & 90.16 & 13.83 & 1.17 & \textbf{59.00} & \textbf{35.90}
\\
    & ClimaX & 96.72 & 10.22 & 0.12 & 24.87 & 9.47 & 93.75 & 5.42 & 0.06 & 15.64 & 8.92
\\\midrule
\multirow{8}{*}{Fully finetune} & U-Net& 99.18& 72.77& 10.74& \textbf{81.55}& \textbf{35.24}& 98.37& 57.22& 5.78& \textbf{73.91}& \textbf{45.60}
\\
    &SegFormer& 99.23 & 71.53 & 12.86 & 75.58 & 33.73 & 98.47 & 55.70 & 6.91 & 64.24 & 40.72
    \\
    &ConvNeXt& \textbf{99.33} & \textbf{73.70} & \textbf{16.87}&  75.37 & 34.14 & \textbf{98.67} & \textbf{58.38} & \textbf{9.22} & 63.79 & 40.89
    \\
    &SatMAE& 88.43 & 10.42 & 1.01 & 59.37 & 24.38 & 79.55 & 5.50 & 0.51 & 45.89 & 27.54
    \\
    &Prithvi& 96.03 & 33.11 & 1.45 & 58.38 & 24.45 & 92.46 & 19.99 & 0.73 & 43.70 & 26.66
    \\
    &Prithvi-2 & 98.86 & 53.68 & 5.79 & 59.76 & 25.79 & 97.75 & 36.69 & 2.99 & 44.52 & 27.91
    \\
    &DOFA*& 94.15 & 20.60 & 2.12 & 70.24 & 28.21 & 88.95 & 11.49 & 1.07 & 57.92 & 34.14
    \\
    &ClimaX & 97.97 & 43.93 & 0.40 & 64.25 & 24.59 & 96.03 & 28.43 & 0.20 & 50.61 & 28.93
    \\\midrule
      
    \multirow{8}{*}{Random init.}& U-Net& \textbf{99.28} & \textbf{74.46} & 0.00 & \textbf{79.21} & 29.97 & \textbf{98.57} & \textbf{59.31} & 0.00 & \textbf{71.97} & 40.85
\\
    &SegFormer& 99.16 & 71.44 & 4.57 & 76.54 & 31.02& 98.34 & 55.58 & 2.37 & 66.83 & 39.56
    \\
    &ConvNeXt &99.24& 71.32& \textbf{13.86} & 76.55 &\textbf{33.74}& 98.48 &55.44& 7.45& 65.81& \textbf{41.26}
    \\
    &SatMAE & 82.01 & 8.73 & 0.71 & 69.45 & 27.56 & 69.66 & 4.57 & 0.36 & 57.27 & 33.49
 \\
    &Prithvi & 97.23 & 37.54 & 2.65 & 54.84 & 25.41 & 94.64 & 23.19 & 1.35 & 39.03 & 26.04
\\
    &Prithvi-2 & 98.93& 53.74 & 7.27 & 60.36 & 25.51 & 97.88 & 36.75 & 3.79 & 45.34 & 27.85
    \\
    &DOFA*& 93.15 & 19.11 & 2.29 & 72.36 & 29.37 & 87.19 & 10.57 & 1.16 & 60.80 & 36.06
    \\
    &ClimaX & 98.90 & 49.70 & 0.08 & 51.94 & 19.69 & 97.83 & 33.14 & 0.04 & 37.05 &	21.06
 \\ 
  
    \bottomrule
    \end{tabular}}
    \caption{\small Flood: F1 and IoU performance of different models. The F1 score and IoU are calculated pixel-wise over the test set for open-flooded and urban-flooded areas. The mF1 and mIoU are calculated pixel-wise over the test set, averaged for each class (macro) and weighted averaged accounting for label imbalance (weighted). The background (non-flooded area) is ignored when computing mIoU and mF1. The best scores are shown in bold.}
    \label{tab:flood_frozenbody}
\end{table}

\subsubsection{Summary}
In summary, the pre-trained weights can accelerate convergence and improve the performance of downstream tasks. By leveraging relevant pretraining data, FMs show generalizability and offer potential solutions to disaster management.

Most Earth-related phenomena involve complex temporal dynamics. Future research should prioritize models with advanced temporal processing capabilities to effectively track changes over time.

While typical vision datasets, Earth Observation datasets, and Weather \& Climate datasets share some similarities, they also present unique challenges. W\&C data often exhibit strong geographical dependencies and specific physical properties. The interactions among variables across broad spatial ranges are deeply rooted in the many atmospheric motion processes. RS imagery can span multiple spectral bands and vary in spatial and temporal resolution. Their data quality is also influenced by sensing conditions, such as illumination and cloud coverage. These characteristics necessitate specialized considerations during model design. Embedding geographical locations, incorporating spectral information, and accounting for ground sampling distance have been shown to improve model performance. For instance, Prithvi-2, compared to Prithvi, incorporates location information and demonstrates superior performance. Future models should go deeper into the spectral characteristics and physical properties of the variables. For instance, the same C-band radar signal can exhibit varying characteristics depending on the polarization mode (e.g., horizontal (HH), vertical (VV), or cross-polarizations (HV or VH)). Ignoring these modes can lead to incomplete or inaccurate interpretations of the data. Therefore, future model designs should consider the sensing characteristics and physical properties to fully capture the complexity of the data.

However, such tailored designs pose challenges in creating unified FMs capable of addressing all tasks. Despite these challenges, our experiments underscore the promise of FMs in offering generalized solutions.

Another critical area for improvement is handling data imbalance, as most models struggle to capture extreme values and minority classes, even after fine-tuning. Developing effective strategies to address this issue remains a key priority for future research.

When selecting a model, end users should carefully evaluate the alignment between the pretrained and downstream data sources. Ideally, the two datasets should align closely in terms of spatial, temporal coverage, resolution, and spectral variables to maximize feature transfer and performance. However, this is often not the case. For developers, pretraining on extensive and diverse datasets can help bridge these gaps and better support end users.

\end{document}